\documentclass[11pt]{article}
\usepackage{pdflscape}
\usepackage[margin=1in]{geometry}
\usepackage[utf8]{inputenc}
\usepackage[T1]{fontenc}
\usepackage{lmodern}
\usepackage{microtype}
\usepackage{amsmath,amssymb,amsfonts}
\usepackage{booktabs}
\usepackage{array}
\usepackage{longtable}
\usepackage{float}
\usepackage{xcolor}
\usepackage{graphicx}
\usepackage[round,authoryear]{natbib}
\usepackage{hyperref}
\usepackage{algorithm}
\usepackage{algpseudocode}
\usepackage{longtable}
\usepackage{booktabs}
\usepackage{pdflscape}
\usepackage{tabularx}
\usepackage{ragged2e}
\usepackage{amsthm}
\newtheorem{assumption}{Assumption}
\newtheorem{proposition}{Proposition}
\newcolumntype{Y}{>{\RaggedRight\arraybackslash}X}
\hypersetup{colorlinks=true,linkcolor=blue,citecolor=blue,urlcolor=blue}

\renewcommand{\arraystretch}{1.15}

\title{Forecasting Medium-Horizon Alzheimer's Disease Progression: Residual Gap-Aware Transformers for 24-Month CDR-SB Change from ADNI Clinical and Biomarker Histories}
\author{
Ran Tong\textsuperscript{1,*,$\dagger$},
Tong Wang\textsuperscript{2,*,$\dagger$},
Lanruo Wang\textsuperscript{3},
Yin Ni\textsuperscript{4}
\\[0.8em]
\small \textsuperscript{1}Department of Mathematical Sciences, University of Texas at Dallas, Richardson, TX 75080, United States\\
\small \textsuperscript{2}Department of Plant Science and Landscape Architecture, University of Connecticut, Storrs, CT, United States\\
\small \textsuperscript{3}Naveen Jindal School of Management, University of Texas at Dallas, Richardson, TX 75080, United States\\
\small \textsuperscript{4}Zhejiang Provincial People's Hospital, Zhejiang, China\\[0.5em]
\small \textsuperscript{*}These authors contributed equally.\\
\small \textsuperscript{$\dagger$}Corresponding authors.
}
\date{}

\begin{document}
\maketitle

\begin{abstract}
Medium-horizon Alzheimer's disease progression prediction is difficult because future clinical scores can remain strongly tied to baseline severity, while longitudinal biomarker histories are irregular and incompletely observed. We develop an anchor-based analysis of 24-month Clinical Dementia Rating Sum of Boxes (CDR-SB) change using harmonized tables derived from the Alzheimer's Disease Neuroimaging Initiative (ADNI). Each labeled sample is anchored at a mild cognitive impairment visit, uses only clinical and biomarker history observed at or before that anchor, and defines the response as CDR-SB at the future visit closest to 24 months within an 18--30 month window minus anchor CDR-SB. The analytic cohort contains 2{,}600 labeled anchors from 858 participants and 7{,}276 longitudinal rows, with cognition, function, demographics, diagnosis, APOE4 allele count (number of apolipoprotein E epsilon 4 alleles), structural magnetic resonance imaging summaries, and cerebrospinal fluid biomarkers aligned by actual visit date.

We propose a residual gap-aware transformer that combines a mixed-effects statistical reference with transformer-based residual learning from pre-anchor longitudinal clinical and biomarker histories. The model uses participant-level random intercepts in the mixed-effects reference, observation-level triplet tokenization for irregular histories, and a learned nonnegative time-gap penalty inside self-attention. The final prediction is the sum of the mixed-effects fixed-effect prediction and the learned transformer residual. We compare the proposed model with a Bayesian-information-criterion-selected linear mixed-effects baseline, GRU-D, and STraTS under repeated participant-level train--test splits. Across five participant-level random seeds, the proposed model achieves the best mean test performance across all reported metrics, reducing MSE by 13.1\% and increasing prediction--observation correlation by 26.4\% relative to the mixed-effects baseline. It also improves over both GRU-D and STraTS in mean error and correlation. These results show that statistical anchoring and gap-aware residual learning provide a useful structure for medium-horizon Alzheimer's disease progression prediction from longitudinal clinical and biomarker ADNI histories.
\end{abstract}

\section{Introduction}

Alzheimer's disease progression is a longitudinal process. Cognitive status, functional impairment, diagnosis, and biomarker burden change over repeated visits, often under irregular follow-up and incomplete biomarker coverage. This structure makes the Alzheimer's Disease Neuroimaging Initiative (ADNI) an important resource for studying disease progression rather than only cross-sectional diagnosis. In a broad review of ADNI publications, \citet{weiner2017adni} showed that ADNI had become a major platform for biomarker validation and longitudinal disease characterization. \citet{veitch2024adni} documented that later ADNI studies continued to support longitudinal modeling, biomarker research, and treatment-era methodological development. Looking across two decades of work, \citet{okonkwo2025adni} described ADNI as a mature infrastructure for prognosis research, biomarker development, and cross-cohort translation. These reviews place ADNI at the center of modern longitudinal Alzheimer's disease research.

Recent work in Alzheimer's disease prediction has also moved toward future-oriented progression modeling.\citet{marinescu2021tadpole} made this direction explicit through the TADPOLE challenge, which organized prediction around future diagnosis and future clinical measurements from ADNI histories. \citet{nguyen2020deep} showed that recurrent neural networks can forecast future diagnosis, cognition, and ventricular measures from longitudinal ADNI data. \citet{alolaimat2023ppad} focused more directly on progression from mild cognitive impairment by developing deep models for future Alzheimer's disease progression. \citet{zhang2024longitudinal} used longitudinal multi-source data to identify disease-related progression patterns associated with prognosis. \citet{lee2024multimodal} developed a machine learning framework for predicting dementia conversion. These studies show that future-oriented Alzheimer's disease prediction is feasible and clinically relevant, while also showing that the target definition and evaluation design strongly shape what a model is being asked to learn.

A central difficulty is that reported predictive gains can be hard to interpret when cohort construction and outcome definition differ across studies. Outcome horizon, anchor state, follow-up window, biomarker handling, and validation protocol all affect the meaning of a model comparison. \citet{grueso2021systematic} documented substantial variation in inclusion criteria, feature construction, and validation among studies of progression from mild cognitive impairment to Alzheimer's disease dementia. \citet{ahmadzadeh2023neuroimaging} reached a similar conclusion for neuroimaging-based transition prediction and emphasized continuing methodological unevenness. \citet{singh2024review} described forecasting-oriented transition studies as a rapidly growing area whose target definitions and evaluation strategies still differ widely. \citet{kumar2021systematic} reviewed clinical-data machine learning studies and highlighted heterogeneity in data sources, tasks, and validation settings. \citet{malik2024review} extended the same concern to the broader Alzheimer's disease prediction literature. These reviews motivate a study design in which the anchor definition, prediction horizon, endpoint, covariate history, and participant-level evaluation rule are stated before model comparison.

The endpoint is especially important. Prior work has shown that the Clinical Dementia Rating Sum of Boxes (CDR-SB) is a meaningful longitudinal outcome in Alzheimer's disease. \citet{cedarbaum2013cdrsb} argued that CDR-SB is a suitable primary outcome because it combines cognitive and functional decline. \citet{williams2013cdrsb} showed that CDR-SB tracks Alzheimer's disease progression over time and carries clinically meaningful longitudinal information. \citet{andrews2019mcid} studied clinically meaningful change in Alzheimer's disease outcome measures, including CDR-SB. \citet{jamalian2023cdrsb} modeled longitudinal CDR-SB trajectories using clinical trial and ADNI data. These studies make CDR-SB a natural outcome family for progression modeling. The remaining design question is which version of CDR-SB gives the clearest medium-horizon progression target. In the cohort used here, raw 24-month CDR-SB remains strongly associated with anchor CDR-SB, whereas 24-month CDR-SB change is much less explained by the anchor value alone. This empirical contrast motivates the primary response in the present study. A change score places the prediction target on worsening after the anchor rather than on baseline disease burden carried into the future score.

Clinical and biomarker history creates a second practical challenge. Structural magnetic resonance imaging and cerebrospinal fluid biomarkers are informative for Alzheimer's disease progression, but they are often unavailable at the exact anchor visit. \citet{lee2019multimodal} showed that combining different data sources can improve Alzheimer's disease progression prediction. \citet{ding2023longitudinal} further showed that longitudinal and multi-source information improves prediction of progression from mild cognitive impairment to Alzheimer's disease. \citet{zhang2024longitudinal} used longitudinal multi-source data to study disease-related progression patterns, and \citet{lee2024multimodal} used multiple data sources for dementia conversion prediction. These studies support the use of clinical and biomarker histories. At the same time, a strict same-visit complete-case rule can sharply reduce the usable cohort. For a medium-horizon longitudinal analysis, historically observed biomarker information remains clinically relevant, and its recency should be represented rather than ignored.

A third challenge is the choice of statistical reference. In an anchor-based longitudinal analysis, the same participant can contribute multiple eligible mild cognitive impairment anchors. This creates within-subject dependence that a simple anchor-level regression does not represent. A mixed-effects model with participant-level random intercepts is therefore a more appropriate statistical comparator for this study design. The proposed neural model is also built around this principle: it learns a residual component beyond the mixed-effects fixed-effect prediction, rather than replacing the longitudinal statistical reference with a black-box sequence model.

This study is built from these considerations. We construct a participant-level, anchor-based analysis of 24-month CDR-SB change using harmonized longitudinal tables derived from ADNI. Each labeled sample is anchored at a visit where the participant is diagnosed with mild cognitive impairment and includes only information observed up to that visit, including repeated clinical scores and available biomarker history. The outcome is defined at the future visit closest to 24 months within an 18--30 month window. To evaluate models in a way that respects repeated measures, we compare the proposed model against a mixed-effects statistical baseline with participant random intercepts, together with recurrent and transformer-based comparators for irregular clinical time series.

The contribution of the paper is threefold. First, it defines a clinically interpretable medium-horizon progression analysis based on 24-month CDR-SB change, with explicit anchor selection, follow-up window construction, biomarker-history handling, and participant-level evaluation. Second, it uses a mixed-effects repeated-measures model as the statistical reference, which makes the comparison more appropriate for repeated anchors from the same participant. Third, it proposes a residual gap-aware transformer that combines the mixed-effects fixed-effect prediction with a transformer residual learned from irregular pre-anchor clinical and biomarker histories, using a nonnegative time-gap penalty inside self-attention. This design links the prediction task, statistical reference, and neural architecture to the same longitudinal data structure. Under this shared quantitative CDR-SB-change analysis, the proposed model achieves the best repeated-seed mean performance among the compared model families.

\section{Related Work}

\subsection{Longitudinal Alzheimer's disease forecasting and progression targets}

Recent Alzheimer's disease prediction studies have increasingly moved from current-state classification toward explicit forecasting of future disease status and future clinical outcomes. \citet{marinescu2021tadpole} played an important role in this shift by organizing the TADPOLE challenge around future diagnosis and future clinical measurements from ADNI histories. That work helped clarify that, once the prediction target moves into the future, cohort definition, horizon choice, and response construction become part of the scientific problem.

\citet{nguyen2020deep} showed that recurrent neural networks can use longitudinal ADNI histories to forecast future diagnosis, cognition, and ventricular measures. Their work demonstrated that repeated observations contain predictive information beyond a single anchor visit and that neural sequence models can use that structure. Their study considered several future targets, which helped establish feasibility while leaving room for a more focused question about how to define one clinically interpretable medium-horizon progression response for direct model comparison.

\citet{alolaimat2023ppad} moved closer to the clinical setting considered here by focusing on progression from mild cognitive impairment to Alzheimer's disease over future visits. That work placed MCI progression directly at the center of model development and showed that longitudinal deep models can be trained for future progression. \citet{zhang2024longitudinal} used longitudinal multi-source data to identify disease-related progression patterns associated with prognosis. \citet{lee2024multimodal} developed a machine learning framework aimed at dementia conversion prediction. \citet{ding2023longitudinal} showed that longitudinal and multi-source information improves prediction of progression from MCI to Alzheimer's disease relative to more limited formulations. Together, these studies show that longitudinal prediction is feasible and clinically relevant, while also showing that the exact target definition remains central.

Prior reviews reinforce this point. \citet{grueso2021systematic} documented variation in inclusion criteria, feature construction, and validation protocols among studies of progression from mild cognitive impairment to Alzheimer's disease dementia. \citet{ahmadzadeh2023neuroimaging} reached a similar conclusion for neuroimaging-based transition prediction. \citet{singh2024review} described forecasting-oriented transition studies as a growing area whose target definitions and evaluation strategies still differ substantially. \citet{kumar2021systematic} and \citet{malik2024review} extended the same concern to broader clinical-data and machine-learning studies of Alzheimer's disease prediction. These reviews matter for the present paper because model performance is often tied to cohort restriction, outcome definition, and validation design. The present study addresses this issue by specifying the horizon, anchor state, response, covariate history, and participant-level split rule before comparing model families.

Quantitative CDR-SB-change outcomes are important in this setting. CDR-SB is an ordered composite clinical scale, but its change over time is widely used to summarize clinical worsening. \citet{cedarbaum2013cdrsb} examined the psychometric properties of CDR-SB in ADNI and argued that it is a strong candidate primary outcome because it reflects both cognitive and functional decline. \citet{williams2013cdrsb} showed that CDR-SB tracks disease progression over time. \citet{andrews2019mcid} studied minimal clinically important differences in Alzheimer's disease outcome measures and gave clinical meaning to changes on scales such as CDR-SB. \citet{jamalian2023cdrsb} modeled longitudinal CDR-SB trajectories using clinical trial and ADNI data. These papers support 24-month CDR-SB change as a meaningful quantitative progression endpoint.

\subsection{Classical longitudinal statistical models and clinically interpretable comparators}

Classical longitudinal modeling provides the statistical basis for the comparator used in this study. In biomedical research, repeated-measures models remain a standard way to describe within-subject dependence, separate population-level and subject-level structure, and support interpretable effect assessment. \citet{laird1982random} gave the foundational random-effects formulation for longitudinal data. \citet{verbeke2000linear} developed the linear mixed-model framework more fully for repeated continuous outcomes. \citet{fitzmaurice2011longitudinal} remains a standard reference for practical longitudinal-data analysis. These works matter directly for the present study because the same participant can contribute multiple eligible anchors, which creates within-subject dependence.

This point affects the choice of comparator. In a repeated-anchor progression analysis, the statistical baseline should reflect the repeated-measures structure of the data. A purely anchor-level regression can serve as a descriptive reference, while a mixed-effects model is the more appropriate longitudinal comparator. This is why the present study uses a mixed-effects baseline with participant random intercepts rather than a simpler cross-sectional regression reference.

The classical longitudinal modeling tradition also connects with Alzheimer's disease progression modeling itself. \citet{jedynak2012computational} proposed a computational disease progression score for the ADNI cohort and treated progression as a latent longitudinal process rather than as a sequence of disconnected classifications. \citet{jamalian2023cdrsb} modeled longitudinal CDR-SB trajectories directly, again emphasizing that progression can be represented as an evolving process rather than only as a future label. A related statistical direction is joint modeling of longitudinal and event data, for which \citet{rizopoulos2012joint} provides the standard treatment. The current paper uses a fixed-horizon change score, while the joint-modeling framework reflects the broader statistical principle that repeated longitudinal measurements are informative about future clinical outcomes.

\subsection{Irregular clinical sequence models and transformer-based time-series methods}

The machine learning work most relevant to the proposed model comes from irregular clinical sequence modeling. \citet{baytas2017tlstm} introduced T-LSTM and showed that elapsed time can be incorporated directly into recurrent dynamics when observations arrive at irregular intervals. This work is relevant here because it treats irregular timing as a structural feature of the sequence.

\citet{che2018grud} developed GRU-D, a recurrent model for multivariate clinical time series with missing values. Its main contribution was to treat both elapsed time and missingness as informative. GRU-D allows the model to learn how stale information decays and how missingness patterns carry signal. That idea is directly relevant to ADNI-style clinical and biomarker histories, where measurements are observed irregularly and biomarker availability changes across visits. GRU-D is therefore one of the most appropriate non-transformer comparators for the data structure used in this paper.

\citet{alolaimat2024tarnn} continued this line of work by combining time-awareness with attention in irregular electronic health record sequences. That paper is useful here because it reinforces the idea that temporal distance should affect interaction strength explicitly. This idea is related to the gap-aware attention term used in the proposed model, although the present paper applies it to a fixed-horizon Alzheimer's disease progression analysis and combines it with a mixed-effects residual structure.

Transformer-based sequence modeling forms the other main methodological branch. \citet{vaswani2017attention} introduced the self-attention architecture underlying modern transformer models. For irregular clinical time series, visits and variables are often observed on different schedules, so a dense regular-grid representation can be poorly matched to the data. \citet{tipirneni2021strats} addressed this problem through STraTS, which represents multivariate clinical history as observation-level triplets rather than as a fully aligned matrix. STraTS is therefore an important comparator in the present study because it shares the triplet-style representation used by the proposed model.

More general transformer work for multivariate time series has also helped define the design space. \citet{zerveas2021transformer} developed a transformer-based framework for multivariate time-series representation learning and showed that transformer models can be adapted beyond language data. The question for the present setting is which structure makes attention better matched to irregular, clinically organized histories.

The proposed method enters at that point. It keeps the observation-level triplet representation associated with STraTS, preserves a mixed-effects statistical reference, and modifies attention through an explicit penalty on temporal distance. The model should therefore be read as a structured statistical-neural model for irregular longitudinal progression prediction. Its main design choices correspond to the data features emphasized throughout this section: repeated anchors within participant, irregular measurement timing, incomplete biomarker coverage, and a fixed-horizon quantitative progression response.

\section{Methodology: Cohort Construction, Outcome Definition, and Longitudinal Data Structure}

This section describes the analytic construction used before model fitting. We define the study cohort, the anchor and follow-up rules, the 24-month CDR-SB-change response, and the clinical and biomarker covariate structure. These steps are part of the methodology because they determine the prediction target, the information available at each anchor, and the participant-level structure used in the subsequent model comparison.

\subsection{Study cohort and follow-up design}

All analyses use harmonized longitudinal tables derived from the Alzheimer's Disease Neuroimaging Initiative (ADNI). The analytic dataset is built by aligning participant identifier, visit code, and actual visit date across diagnosis records, cognitive and functional assessments, demographics, APOE4 allele count (number of apolipoprotein E epsilon 4 alleles), structural magnetic resonance imaging regional volume tables, and cerebrospinal fluid biomarker tables. Actual visit date is used because nominal visit labels alone do not resolve missed visits, asynchronous modality collection, or irregular follow-up. In a medium-horizon progression study, timing determines which observations belong to the available history, which future visits are eligible for outcome assignment, and how informative historical biomarker measurements remain at the anchor.

The study is anchored at visits where the participant is diagnosed with mild cognitive impairment (MCI). This anchor state is clinically useful because future worsening remains plausible and clinically important, yet there is still substantial heterogeneity in cognitive burden, functional status, and biomarker pattern across subjects. For each anchor visit, the study searches for future follow-up visits occurring between 18 and 30 months after the anchor. Among those eligible visits, the visit closest to 24 months is selected as the outcome visit. This rule yields a clinically interpretable medium-horizon design while remaining feasible under irregular ADNI follow-up.

The primary CDR-SB-change cohort retains anchor visits for which both anchor and selected future CDR-SB values are observed. Under this rule, the resulting analytic cohort contains 2{,}600 labeled anchor visits from 858 participants and 7{,}276 longitudinal rows. The median target gap is 24.11 months. Same-visit structural magnetic resonance imaging availability is 17.1\%, whereas any-prior structural magnetic resonance imaging availability rises to 36.4\%. The corresponding numbers for cerebrospinal fluid are 22.0\% for same-visit availability and 60.5\% for any-prior availability. These quantities show why strict same-visit biomarker coverage would be poorly matched to the data structure. Historical biomarker information is much more broadly available than same-visit biomarker information alone. Table~\ref{tab:cohort-profile} summarizes the labeled cohort size, repeated-anchor structure, target timing, and same-visit versus any-prior biomarker availability used in the primary 24-month CDR-SB-change analysis.

\begin{table}[t]
    \centering
    \caption{Profile of the primary 24-month CDR-SB-change cohort derived from ADNI. Percentages are relative to labeled anchor visits.}
    \label{tab:cohort-profile}
    \small
    \setlength{\tabcolsep}{7pt}
    \renewcommand{\arraystretch}{1.25}
    \begin{tabular}{p{0.48\linewidth}p{0.14\linewidth}p{0.26\linewidth}}
        \toprule
        Cohort characteristic & Value & Interpretation \\
        \midrule
        Labeled anchor visits with observed 24-month CDR-SB change & 2{,}600 & Effective sample size for the main CDR-SB-change analysis \\
        Participants & 858 & Repeated anchors arise within participant and create within-subject dependence \\
        Total longitudinal rows & 7{,}276 & Historical observations available before anchor construction \\
        Median target gap (months) & 24.11 & Outcome assignment is centered near the intended 24-month horizon \\
        Same-visit structural magnetic resonance imaging availability & 17.1\% & Same-visit imaging availability is limited \\
        Any-prior structural magnetic resonance imaging availability & 36.4\% & Prior imaging history substantially broadens usable biomarker coverage \\
        Same-visit cerebrospinal fluid availability & 22.0\% & Same-visit cerebrospinal fluid availability remains limited at the anchor \\
        Any-prior cerebrospinal fluid availability & 60.5\% & Historical cerebrospinal fluid information contributes meaningful additional cohort coverage \\
        \bottomrule
    \end{tabular}
\end{table}

\subsection{Primary outcome and statistical rationale}

The primary outcome is the 24-month Clinical Dementia Rating Sum of Boxes (CDR-SB) change,
\begin{equation}
    \Delta \mathrm{CDR\mbox{-}SB}_{i,24m}
    =
    \mathrm{CDR\mbox{-}SB}_{i,24m}
    -
    \mathrm{CDR\mbox{-}SB}_{i,0},
\end{equation}
whenever both values are observed. Throughout the paper, this quantity is referred to as the \emph{24-month CDR-SB change}. This endpoint measures worsening magnitude over a clinically interpretable follow-up interval. Because CDR-SB is an ordered composite clinical scale, the response is interpreted as a quantitative change score rather than as a strictly continuous biological measurement.

The choice of CDR-SB is supported by prior clinical and longitudinal work. \citet{cedarbaum2013cdrsb} argued that CDR-SB is a suitable primary outcome for Alzheimer's disease trials because it combines cognitive and functional decline in a single clinical measure. \citet{williams2013cdrsb} showed that CDR-SB scores track Alzheimer's disease progression over time, supporting their use in longitudinal progression analysis. \citet{andrews2019mcid} studied clinically meaningful change in Alzheimer's disease outcome measures, including CDR-SB, which helps connect score changes with clinical interpretation. \citet{jamalian2023cdrsb} modeled longitudinal CDR-SB trajectories using clinical trial and ADNI data, further supporting CDR-SB as a progression endpoint in longitudinal modeling.

The key design question is which version of CDR-SB gives the clearest medium-horizon progression target. A natural alternative is the raw future CDR-SB level at the selected 24-month visit. That quantity is clinically interpretable, but it remains strongly tied to the anchor value. In the present cohort, current CDR-SB alone has correlation 0.667 with raw 24-month CDR-SB and explains 44.4\% of its variance. By contrast, current CDR-SB has correlation only 0.175 with 24-month CDR-SB change and explains 3.1\% of its variance. This contrast is central to the design of the study. A raw future score still carries substantial information about where the subject started. A change score moves the target toward worsening after the anchor.

Each labeled sample therefore represents a clinically interpretable question asked at an MCI visit: given the subject's repeated clinical history and available biomarker history up to this point, how much worsening is recorded approximately two years later? That question is narrower than a generic future-state forecast and better aligned with the quantitative progression problem studied in this paper.

\subsection{Covariate structure, biomarker history, and derived variables}

The covariate structure is organized around the same principle as the outcome: the data used for prediction should reflect what is observed before the anchor and should preserve the longitudinal clinical and biomarker structure of the cohort. The analysis therefore includes repeated clinical measures, anchor-level characteristics, historical biomarker summaries, and derived variables that describe missingness and timing.

Longitudinal clinical variables include CDR-SB, MMSE, ADAS13, and FAQ measured over repeated visits before the anchor. These variables provide the densest part of the subject history and capture both disease burden and its temporal evolution. Structural magnetic resonance imaging summaries include regional and global measurements such as whole brain, ventricles, hippocampus, entorhinal cortex, fusiform gyrus, middle temporal cortex, and inferior temporal cortex whenever observed before the anchor. Cerebrospinal fluid history includes amyloid-$\beta$, tau, and phosphorylated tau when available. These biomarker histories are retained because prior values remain scientifically informative even when they are not observed at the exact anchor visit.

Anchor-level and slowly varying variables include age, sex, education, APOE4 allele count, baseline diagnostic context, months from first visit, and exact target-gap months. These quantities provide stable clinical context and disease-history information that complement the repeated measures. They are also the natural covariates for the mixed-effects reference model used later in the paper.

Derived variables are used to represent the irregular structure of follow-up explicitly. For statistical modeling, the analysis includes missingness indicators and modality-recency variables that record how long it has been since structural magnetic resonance imaging or cerebrospinal fluid was last observed. For sequence models, the same history is represented through observation-level triplets and elapsed-time tensors. This construction keeps timing and modality availability explicit rather than hiding them inside strict complete-case filtering. Table~\ref{tab:variables-expanded} lists the clinical, biomarker, timing, missingness, sequence-format, and response variables used to construct the statistical and neural model inputs.

\clearpage
\begin{landscape}

\begin{table}[p]
\centering
\caption{Covariate and outcome structure for the 24-month CDR-SB-change analysis. Variables are grouped into longitudinal clinical predictors, longitudinal biomarker predictors, anchor-level predictors, derived timing and missingness variables, sequence-format variables, and the response.}
\label{tab:variables-expanded}

\scriptsize
\setlength{\tabcolsep}{4pt}
\renewcommand{\arraystretch}{1.12}

\begin{tabularx}{\linewidth}{
    >{\RaggedRight\arraybackslash}p{0.17\linewidth}
    Y
    >{\RaggedRight\arraybackslash}p{0.18\linewidth}
    Y
}
\toprule
Block & Variables included & Temporal status & Role in the analysis \\
\midrule

\multicolumn{4}{l}{\textit{Longitudinal clinical and functional predictors}} \\
Clinical severity &
CDR-SB, MMSE, ADAS13 &
Repeated pre-anchor measurements &
Capture cognitive burden and its pre-anchor trajectory. These variables provide the main longitudinal clinical signal. \\

Functional status &
FAQ &
Repeated pre-anchor measurements &
Captures functional impairment before the anchor and complements cognitive scores. \\
\midrule

\multicolumn{4}{l}{\textit{Longitudinal biomarker predictors}} \\
Structural magnetic resonance imaging history &
Whole brain, ventricles, hippocampus, entorhinal cortex, fusiform gyrus, middle temporal cortex, inferior temporal cortex &
Observed at or before anchor when available &
Summarizes regional and global structural neurodegeneration. Prior values are retained when same-visit magnetic resonance imaging is unavailable. \\

Cerebrospinal fluid history &
Amyloid-$\beta$, tau, phosphorylated tau &
Observed at or before anchor when available &
Represents molecular pathology before the anchor and complements clinical and structural magnetic resonance imaging information. \\
\midrule

\multicolumn{4}{l}{\textit{Anchor-level and slowly varying predictors}} \\
Demographic variables &
Age, sex, education &
Anchor-level or effectively fixed &
Provide background clinical adjustment and account for demographic heterogeneity. \\

Genetic variable &
APOE4 allele count &
Fixed &
Encodes inherited risk information related to Alzheimer's disease progression. \\

Anchor diagnostic context &
MCI anchor indicator and available diagnosis coding &
Anchor-level &
Records the clinical state from which future worsening is modeled and preserves the diagnostic coding used during cohort construction. \\

Disease-history timing &
Months from first visit, exact target-gap months &
Anchor-level derived quantities &
Encodes duration of observed follow-up history and the exact distance between anchor and outcome visit. \\
\midrule

\multicolumn{4}{l}{\textit{Derived variables for irregular clinical and biomarker follow-up}} \\
History-aware magnetic resonance imaging summaries &
Most recent prior structural magnetic resonance imaging summaries &
Derived from pre-anchor history &
Preserves structural biomarker information when same-visit imaging is unavailable. \\

History-aware cerebrospinal fluid summaries &
Most recent prior amyloid-$\beta$, tau, and phosphorylated tau summaries &
Derived from pre-anchor history &
Preserves molecular biomarker information when same-visit cerebrospinal fluid is unavailable. \\

Missingness indicators &
Variable-specific missingness flags for clinical, structural magnetic resonance imaging, and cerebrospinal fluid features &
Derived from observation pattern &
Records which measurements are observed and allows missing-data patterns to enter the statistical and neural models. \\

Modality recency &
Months since structural magnetic resonance imaging was last observed; months since cerebrospinal fluid was last observed &
Derived from observation timing &
Distinguishes recent biomarker measurements from older carried-forward history. \\
\midrule

\multicolumn{4}{l}{\textit{Sequence-format variables for neural models}} \\
Observation-level triplets &
Observation time, variable identity, and observed value, denoted $(\tau_j,k_j,v_j)$ &
Derived from longitudinal history &
Input representation for STraTS and the proposed gap-aware transformer. \\

Elapsed-time tensors &
Time since last observation by variable &
Derived from longitudinal history &
Input representation for GRU-D and other time-aware recurrent models. \\
\midrule

\multicolumn{4}{l}{\textit{Response}} \\
Primary response &
24-month CDR-SB change &
Future value minus anchor value &
Quantitative regression target for medium-horizon clinical worsening. \\
\bottomrule
\end{tabularx}

\end{table}

\end{landscape}
\clearpage

Table~\ref{tab:analytic-design} summarizes the analytic construction of the study, including the data source, anchor definition, available history, outcome window, response construction, biomarker handling, repeated-anchor structure, and evaluation unit.

\begin{table}[H]
    \centering
    \caption{Analytic construction of the 24-month CDR-SB-change study.}
    \label{tab:analytic-design}
    \small
    \setlength{\tabcolsep}{7pt}
    \renewcommand{\arraystretch}{1.25}
    \begin{tabular}{p{0.24\linewidth}p{0.34\linewidth}p{0.30\linewidth}}
        \toprule
        Component & Definition in this study & Purpose \\
        \midrule
        Data source &
        Harmonized longitudinal tables derived from ADNI &
        Provides repeated clinical, demographic, genetic, structural magnetic resonance imaging, and cerebrospinal fluid information. \\
        \addlinespace[0.25em]
        Anchor visit &
        Visit at which the participant is diagnosed with mild cognitive impairment &
        Defines the clinical state from which future worsening is predicted. \\
        \addlinespace[0.25em]
        Available history &
        All observed information at or before the anchor visit &
        Prevents future information from entering model inputs. \\
        \addlinespace[0.25em]
        Outcome visit &
        Future visit closest to 24 months among visits 18--30 months after the anchor &
        Gives a medium-horizon target while allowing irregular follow-up. \\
        \addlinespace[0.25em]
        Primary response &
        CDR-SB at the selected future visit minus CDR-SB at the anchor &
        Measures clinical worsening after the anchor as a quantitative regression target. \\
        \addlinespace[0.25em]
        Biomarker handling &
        Same-visit biomarker values when available; otherwise most recent prior structural magnetic resonance imaging or cerebrospinal fluid summaries with recency variables &
        Retains informative biomarker history without restricting the analysis to strict same-visit biomarker coverage. \\
        \addlinespace[0.25em]
        Repeated anchors &
        Multiple eligible anchors can arise from the same participant &
        Preserves longitudinal information and motivates participant-level splitting and mixed-effects modeling. \\
        \addlinespace[0.25em]
        Evaluation unit &
        Participant-level train--test split &
        Prevents anchors from the same participant appearing in both training and test sets. \\
        \bottomrule
    \end{tabular}
\end{table}

\section{Proposed Model and Structural Properties}
\label{sec:proposed-method}

This section presents the proposed residual gap-aware transformer for 24-month CDR-SB change. The model combines a mixed-effects statistical reference with a transformer encoder for irregular longitudinal clinical and biomarker histories. The mixed-effects component provides a longitudinal statistical reference based on anchor-level covariates, while the transformer component learns residual trajectory information from the pre-anchor history. The section first defines the model, then states the structural assumptions and propositions used to characterize the residual decomposition and the time-gap attention mechanism. Detailed proofs are provided in Appendix~\ref{app:proofs}.

\begin{figure}[t]
    \centering
    \includegraphics[width=0.95\linewidth]{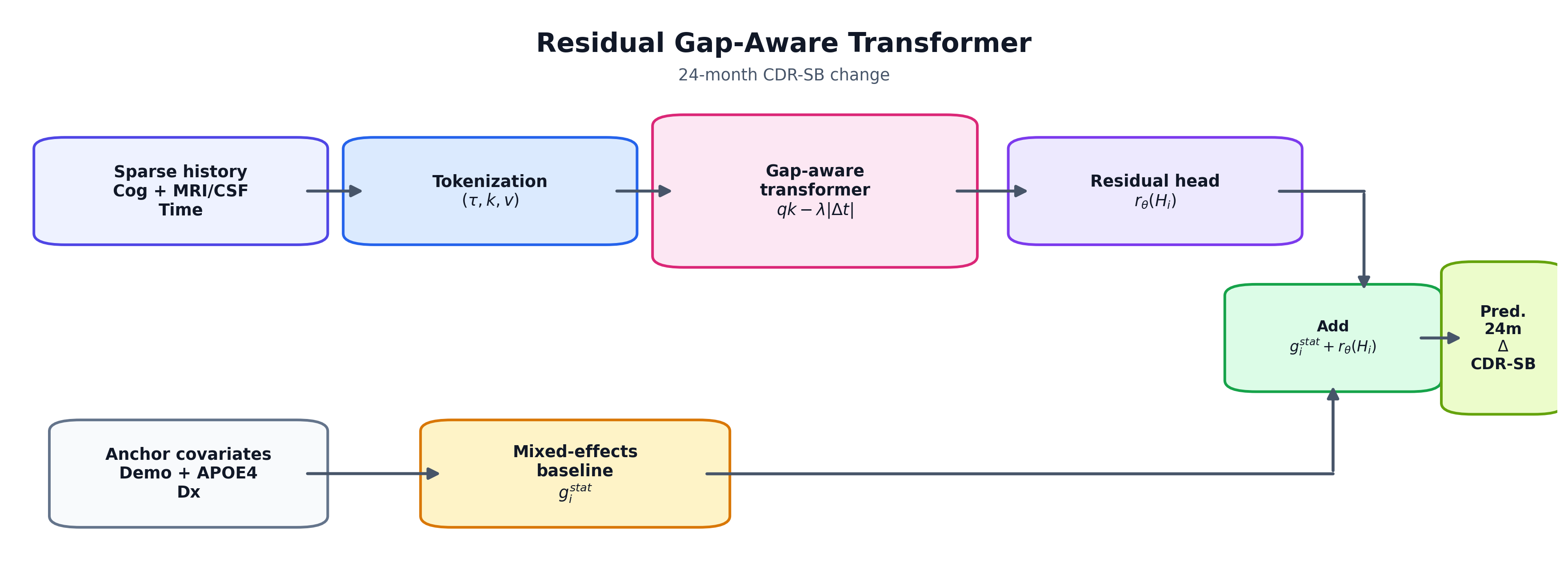}
    \caption{Overview of the proposed residual gap-aware transformer for predicting 24-month CDR-SB change. The upper branch uses anchor-level covariates to obtain the mixed-effects statistical reference prediction \(g_i^{\mathrm{stat}}\). The lower branch encodes pre-anchor longitudinal clinical and biomarker history through observation-level triplet tokenization and a gap-aware transformer, producing the residual prediction \(r_{\theta}(H_i)\). The final prediction is the sum \(g_i^{\mathrm{stat}} + r_{\theta}(H_i)\).}
    \label{fig:model-overview}
\end{figure}

\subsection{Notation and prediction target}

For notational simplicity, this section indexes labeled anchor samples by a single index \(i\). Participant identifiers are used when fitting the mixed-effects reference model and when defining participant-level train--test splits in Section~\ref{sec:experimental-protocol}.

For anchor \(i\), let
\[
H_i=\{(\tau_{ij},k_{ij},v_{ij})\}_{j=1}^{m_i}
\]
denote the longitudinal history observed at or before the anchor. Here \(\tau_{ij}\in\mathbb{R}\) is the observation time measured relative to the anchor, \(k_{ij}\in\{1,\dots,K\}\) is the variable identity, and \(v_{ij}\in\mathbb{R}\) is the observed value. Let \(p_i\) denote the participant identifier, let \(x_i\in\mathbb{R}^{p}\) denote the anchor-level covariate vector used by the mixed-effects reference model, and let
\[
y_i=\Delta \mathrm{CDR\mbox{-}SB}_{i,24m}
\]
be the observed 24-month CDR-SB change.

The mixed-effects reference model is fit on the training set only using participant identifiers, anchor-level covariates, and observed responses. Let
\[
g_i^{\mathrm{stat}}=g_{\hat\beta}(x_i)
\]
denote its fixed-effect prediction for anchor \(i\). In all residual-learning and test-prediction steps, \(g_i^{\mathrm{stat}}\) denotes the marginal fixed-effect prediction from the fitted mixed-effects reference model. Participant-specific random intercepts are used to fit the reference model in the training set, but they are not used as test-subject information.

The proposed model predicts
\begin{equation}
    \hat y_i
    =
    g_i^{\mathrm{stat}} + r_\theta(H_i),
    \label{eq:decomp-main}
\end{equation}
where \(r_\theta(H_i)\) is a residual component learned from the pre-anchor longitudinal history. This decomposition gives the model two components: a statistical reference term and a sequence-based residual term.

Define the residual response
\begin{equation}
    u_i = y_i - g_i^{\mathrm{stat}}.
    \label{eq:residual-target}
\end{equation}
Then fitting the model in \eqref{eq:decomp-main} is equivalent to fitting \(u_i\) using the longitudinal history \(H_i\). This equivalence is formalized in Proposition~\ref{prop:residual-equivalence}.

\subsection{Observation-level longitudinal tokenization}

Each observed triplet \((\tau_{ij},k_{ij},v_{ij})\) is mapped to an initial token embedding
\begin{equation}
    z_{ij}^{(0)}
    =
    e_{\tau}(\tau_{ij}) + e_{k}(k_{ij}) + W_v v_{ij} + b_v,
    \label{eq:triplet-embed}
\end{equation}
where \(e_{\tau}(\cdot)\) is a time embedding, \(e_{k}(\cdot)\) is a variable-identity embedding, and \(W_v v_{ij}+b_v\) embeds the observed measurement value. Let
\[
Z_i^{(0)}
=
\big(z_{i1}^{(0)},\dots,z_{im_i}^{(0)}\big)^\top
\in \mathbb{R}^{m_i\times d}
\]
be the initial token matrix. This representation uses observed events directly and avoids forcing all variables onto a dense common time grid.

\subsection{Gap-aware self-attention}

The encoder follows the standard multi-head self-attention structure, with a time-gap penalty added to the pre-softmax attention score. For layer \(\ell\) and head \(h\), define
\[
q_{ia}^{(\ell,h)} = W_Q^{(\ell,h)} z_{ia}^{(\ell-1)}, \qquad
k_{ib}^{(\ell,h)} = W_K^{(\ell,h)} z_{ib}^{(\ell-1)}, \qquad
v_{ib}^{(\ell,h)} = W_V^{(\ell,h)} z_{ib}^{(\ell-1)}.
\]
The attention score from query token \(a\) to key token \(b\) is
\begin{equation}
    s_{iab}^{(\ell,h)}
    =
    \frac{\langle q_{ia}^{(\ell,h)}, k_{ib}^{(\ell,h)}\rangle}{\sqrt{d_h}}
    -
    \lambda_{\ell,h}\,|\tau_{ia}-\tau_{ib}|,
    \label{eq:gap-score}
\end{equation}
where
\begin{equation}
    \lambda_{\ell,h}=\mathrm{softplus}(\eta_{\ell,h}) \ge 0.
    \label{eq:gap-lambda}
\end{equation}
The corresponding attention weight is
\begin{equation}
    \alpha_{iab}^{(\ell,h)}
    =
    \frac{\exp(s_{iab}^{(\ell,h)})}
    {\sum_{c=1}^{m_i}\exp(s_{iac}^{(\ell,h)})}.
    \label{eq:gap-softmax}
\end{equation}
The output of head \(h\) at token \(a\) is
\begin{equation}
    o_{ia}^{(\ell,h)}
    =
    \sum_{b=1}^{m_i}
    \alpha_{iab}^{(\ell,h)}v_{ib}^{(\ell,h)}.
    \label{eq:head-output}
\end{equation}
The multi-head output is
\[
o_{ia}^{(\ell)}
=
W_O^{(\ell)}
\mathrm{Concat}\big(o_{ia}^{(\ell,1)},\dots,o_{ia}^{(\ell,H)}\big).
\]
The token representation is updated through a residual feed-forward block,
\begin{equation}
    z_{ia}^{(\ell)}
    =
    \mathrm{FFN}^{(\ell)}
    \!\left(
    z_{ia}^{(\ell-1)} + o_{ia}^{(\ell)}
    \right).
    \label{eq:transformer-update}
\end{equation}

The nonnegative parameter \(\lambda_{\ell,h}\) controls the strength of temporal attenuation in layer \(\ell\) and head \(h\). For fixed content similarity, larger temporal distance lowers the pre-softmax score. Propositions~\ref{prop:gap-score} and \ref{prop:gap-softmax} state this property for the score and the resulting softmax weight.

\subsection{Pooling and residual regression head}

After \(L\) transformer layers, the encoded tokens are
\[
Z_i^{(L)}
=
\big(z_{i1}^{(L)},\dots,z_{im_i}^{(L)}\big)^\top .
\]
A learned pooling layer converts the variable-length history into a fixed-dimensional representation. Define
\begin{equation}
    \pi_{ij}
    =
    \frac{\exp\!\big(w_p^\top \tanh(W_p z_{ij}^{(L)}+b_p)\big)}
    {\sum_{c=1}^{m_i}\exp\!\big(w_p^\top \tanh(W_p z_{ic}^{(L)}+b_p)\big)},
    \label{eq:pooling-weight}
\end{equation}
and
\begin{equation}
    h_i
    =
    \sum_{j=1}^{m_i}\pi_{ij}z_{ij}^{(L)}.
    \label{eq:pooled-repr}
\end{equation}
The residual prediction is
\begin{equation}
    r_\theta(H_i)
    =
    w_r^\top h_i + b_r.
    \label{eq:residual-head}
\end{equation}
Combining \eqref{eq:decomp-main} and \eqref{eq:residual-head} gives the final predictor
\begin{equation}
    \hat y_i
    =
    g_i^{\mathrm{stat}} + w_r^\top h_i + b_r.
    \label{eq:final-predictor}
\end{equation}

\subsection{Training objective}

For anchor \(i\), the squared-error loss is
\begin{equation}
    \ell_i(\theta)
    =
    \left(
    y_i-g_i^{\mathrm{stat}}-r_\theta(H_i)
    \right)^2.
    \label{eq:point-loss}
\end{equation}
The empirical objective is
\begin{equation}
    \mathcal{L}_n(\theta)
    =
    \frac{1}{n}
    \sum_{i=1}^n
    \left(
    y_i-g_i^{\mathrm{stat}}-r_\theta(H_i)
    \right)^2.
    \label{eq:empirical-loss}
\end{equation}
Using the residual response \(u_i\) in \eqref{eq:residual-target}, this objective can also be written as
\begin{equation}
    \mathcal{L}_n(\theta)
    =
    \frac{1}{n}
    \sum_{i=1}^n
    \left(
    u_i-r_\theta(H_i)
    \right)^2.
    \label{eq:residual-loss}
\end{equation}
This shows that the transformer component is trained as a residual regression model relative to the fitted mixed-effects reference.

\subsection{Training algorithm}

Algorithm~\ref{alg:proposed} summarizes the training procedure. The mixed-effects reference is fit first on the training anchors only. Its fixed-effect predictions are then used as reference values. The gap-aware transformer is trained under the residual regression objective in \eqref{eq:empirical-loss}.

\begin{algorithm}[H]
\caption{Training the proposed residual gap-aware transformer for 24-month CDR-SB change}
\label{alg:proposed}
\begin{algorithmic}[1]
\Require Training anchors $\{(p_i,H_i,x_i,y_i)\}_{i=1}^n$ with participant identifiers \(p_i\), observation-level histories \(H_i\), anchor-level covariates \(x_i\), and targets \(y_i=\Delta \mathrm{CDR\mbox{-}SB}_{i,24m}\); epochs \(E\); batch size \(B\); learning rate \(\eta\)
\State Fit the training-only mixed-effects reference model on $\{(p_i,x_i,y_i)\}_{i=1}^n$
\State Compute fixed-effect reference predictions \(g_i^{\mathrm{stat}}\) for all training anchors
\State Initialize transformer parameters \(\theta\) and gap-penalty parameters \(\{\eta_{\ell,h}\}\)
\For{\(e=1,\dots,E\)}
    \State Shuffle the training anchors
    \For{each mini-batch \(\mathcal{B}\) of size \(B\)}
        \State Convert each history \(H_i\) in \(\mathcal{B}\) to observation-level triplets \((\tau_j,k_j,v_j)\)
        \State Embed triplets and run the transformer encoder
        \State Compute gap-aware attention scores using
        \[
        s_{ab}^{(\ell,h)}
        =
        \frac{\langle q_a^{(\ell,h)}, k_b^{(\ell,h)}\rangle}{\sqrt{d_h}}
        -
        \mathrm{softplus}(\eta_{\ell,h})|\tau_a-\tau_b|
        \]
        \State Produce residual predictions \(r_\theta(H_i)\) for all \(i\in\mathcal{B}\)
        \State Form final predictions \(\hat y_i = g_i^{\mathrm{stat}} + r_\theta(H_i)\)
        \State Compute mini-batch loss
        \[
        \mathcal{L}_{\mathcal{B}}
        =
        \frac{1}{|\mathcal{B}|}
        \sum_{i\in\mathcal{B}}(\hat y_i-y_i)^2
        \]
        \State Update \(\theta\) and \(\{\eta_{\ell,h}\}\) by backpropagation with learning rate \(\eta\)
    \EndFor
    \State Evaluate validation error and apply early stopping when validation performance stops improving
\EndFor
\State \Return trained model \(\hat y_i = g_i^{\mathrm{stat}} + r_\theta(H_i)\)
\end{algorithmic}
\end{algorithm}

\subsection{Assumptions and structural properties}

The following assumptions state the regularity conditions used to characterize the proposed model. The results are deterministic properties of the empirical objective and the model map. They formalize two features of the architecture: residualization relative to the mixed-effects reference model and temporal attenuation induced by the gap-aware attention score. Complete proofs are provided in Appendix~\ref{app:proofs}.

\begin{assumption}[Finite observed response and finite reference prediction]
\label{assump:finite}
For every training anchor \(i\), the observed response \(y_i\) is finite, and the fitted statistical reference prediction \(g_i^{\mathrm{stat}}\) is finite.
\end{assumption}

\begin{assumption}[Feasible zero residual]
\label{assump:zero}
The residual function class
\[
\mathcal{R}=\{r_\theta:\theta\in\Theta\}
\]
contains the zero function. That is, there exists \(\theta_0\in\Theta\) such that
\[
r_{\theta_0}(H)=0
\]
for every admissible history \(H\).
\end{assumption}

\begin{assumption}[Nonnegative temporal gap penalty]
\label{assump:gap}
For every layer \(\ell\) and head \(h\),
\[
\lambda_{\ell,h}=\mathrm{softplus}(\eta_{\ell,h})\ge 0.
\]
\end{assumption}

\begin{assumption}[Bounded admissible histories and regularity of the residual map]
\label{assump:lipschitz}
The admissible history space \(\mathcal{H}\) is equipped with a metric \(d_{\mathcal{H}}\). Each history has at most \(m_{\max}<\infty\) observed tokens, and observed times and values lie in bounded sets after preprocessing. Variable identities are equipped with the discrete metric. The tokenization map from histories to initial embeddings is Lipschitz on \(\mathcal{H}\). For fixed parameter values, all linear maps in the attention blocks, feed-forward blocks, pooling layer, and residual head have finite operator norms on the admissible domain. The activation functions used in the model are Lipschitz on the relevant bounded range.
\end{assumption}

\begin{proposition}[Residual-risk equivalence]
\label{prop:residual-equivalence}
For fixed statistical reference predictions \(\{g_i^{\mathrm{stat}}\}_{i=1}^n\), minimizing
\[
\mathcal{L}_n(\theta)
=
\frac{1}{n}
\sum_{i=1}^n
\left(
y_i-g_i^{\mathrm{stat}}-r_\theta(H_i)
\right)^2
\]
over \(\theta\) is equivalent to minimizing
\[
\frac{1}{n}
\sum_{i=1}^n
\left(
u_i-r_\theta(H_i)
\right)^2,
\qquad
u_i=y_i-g_i^{\mathrm{stat}}.
\]
\end{proposition}

\begin{proposition}[Feasibility of the statistical reference in the empirical objective]
\label{prop:no-worse}
Under Assumption~\ref{assump:zero},
\begin{equation}
    \inf_{\theta\in\Theta}\mathcal{L}_n(\theta)
    \le
    \frac{1}{n}
    \sum_{i=1}^n
    \left(y_i-g_i^{\mathrm{stat}}\right)^2.
    \label{eq:no-worse}
\end{equation}
\end{proposition}

\begin{proposition}[Temporal monotonicity of the gap-aware score]
\label{prop:gap-score}
Fix a layer \(\ell\), a head \(h\), a query token \(a\), and a key token \(b\). Holding the content term
\[
c_{ab}^{(\ell,h)}
=
\frac{\langle q_a^{(\ell,h)},k_b^{(\ell,h)}\rangle}{\sqrt{d_h}}
\]
fixed, the pre-softmax attention score
\[
s_{ab}^{(\ell,h)}(d)
=
c_{ab}^{(\ell,h)}-\lambda_{\ell,h}d,
\qquad
d=|\tau_a-\tau_b|,
\]
is nonincreasing in \(d\) and strictly decreasing whenever \(\lambda_{\ell,h}>0\).
\end{proposition}

\begin{proposition}[Temporal monotonicity of the softmax attention weight]
\label{prop:gap-softmax}
Fix a query token \(a\), a head \(h\), and all key scores except that of one key token \(b\). Suppose \(\lambda_{\ell,h}>0\) and the content term for token \(b\) is held fixed. Then the corresponding softmax attention weight
\[
\alpha_b(d)
=
\frac{\exp(c_b-\lambda_{\ell,h}d)}
{\sum_{c\neq b}\exp(s_c)+\exp(c_b-\lambda_{\ell,h}d)}
\]
is nonincreasing as \(d=|\tau_a-\tau_b|\) increases. If token \(b\) competes with at least one other key token, the decrease is strict.
\end{proposition}

\begin{proposition}[Stability of the residual predictor]
\label{prop:stability}
Under Assumption~\ref{assump:lipschitz}, the residual predictor \(r_\theta(\cdot)\) is Lipschitz on the admissible history domain. That is, there exists a finite constant \(L_\theta\) such that
\begin{equation}
    |r_\theta(H)-r_\theta(H')|
    \le
    L_\theta d_{\mathcal{H}}(H,H')
    \label{eq:lipschitz-main}
\end{equation}
for any two admissible histories \(H,H'\in\mathcal{H}\).
\end{proposition}

These propositions give the formal role of the two structural components in the model. Propositions~\ref{prop:residual-equivalence} and~\ref{prop:no-worse} show that the transformer is trained as a residual regression model relative to the mixed-effects reference and that the empirical objective contains the statistical reference as a feasible special case. Propositions~\ref{prop:gap-score} and~\ref{prop:gap-softmax} show that the learned nonnegative gap penalty induces temporal attenuation in the attention mechanism. Proposition~\ref{prop:stability} gives a stability property of the residual predictor under boundedness and Lipschitz conditions. These results clarify the structure of the proposed model; they should be interpreted as structural properties rather than as a full finite-sample generalization theory.

\section{Experiments: Comparator Models and Evaluation Protocol}
\label{sec:experimental-protocol}

This section describes the experimental design used for the 24-month CDR-SB-change analysis. All models are evaluated under the same anchor construction, participant-level splitting rule, training--validation--test logic, and test metrics. The comparator set is chosen to cover three modeling families that are relevant to the proposed method: longitudinal mixed-effects modeling, missingness-aware recurrent modeling, and transformer-based modeling for irregular clinical time series.

\subsection{Comparator models}

\paragraph{Linear mixed-effects baseline.}
The first comparator is a linear mixed-effects model. Mixed-effects models are standard for repeated-measures data because they separate population-level fixed effects from participant-level random variation. \citet{laird1982random} introduced the random-effects formulation for longitudinal data, \citet{verbeke2000linear} gave a full treatment of linear mixed models for repeated outcomes, and \citet{fitzmaurice2011longitudinal} emphasized their use for within-subject dependence in applied longitudinal studies.

For anchor \(j\) from participant \(i\), the general linear mixed-effects model is written as
\begin{equation}
    y_{ij}
    =
    x_{ij}^{\top}\beta
    +
    z_{ij}^{\top}b_i
    +
    \varepsilon_{ij},
    \label{eq:lmm-general-section5}
\end{equation}
where
\[
    y_{ij}
    =
    \Delta \mathrm{CDR\mbox{-}SB}_{ij,24m}.
\]
Here \(x_{ij}\) is the fixed-effect covariate vector, \(\beta\) is the fixed-effect coefficient vector, \(z_{ij}\) is the random-effect design vector, \(b_i\) is the participant-specific random-effect vector, and \(\varepsilon_{ij}\) is the residual error. The distributional assumptions are
\begin{equation}
    b_i \sim N(0,D),
    \qquad
    \varepsilon_{ij}\sim N(0,\sigma^2),
    \qquad
    b_i \perp \varepsilon_{ij}.
    \label{eq:lmm-distribution-section5}
\end{equation}
In the implemented baseline, the random-effect structure is a participant-level random intercept. Thus \(z_{ij}=1\), \(b_i=b_{0i}\), and the fitted model is
\begin{equation}
    y_{ij}
    =
    \beta_0
    +
    x_{ij}^{\top}\beta
    +
    b_{0i}
    +
    \varepsilon_{ij},
    \qquad
    b_{0i}\sim N(0,\sigma_b^2).
    \label{eq:lmm-random-intercept-section5}
\end{equation}
This form keeps the repeated-anchor dependence explicit while avoiding a more complex random-slope structure that would be harder to estimate stably under the participant-level train--test split.

The fixed-effect covariates include current cognitive and functional measurements, target-gap information, demographics, APOE4 allele count, anchor diagnostic context, structural magnetic resonance imaging history summaries, cerebrospinal fluid biomarker history summaries, missingness indicators, and modality-recency covariates. Candidate fixed-effect sets are selected within the training partition only by the Bayesian information criterion,
\begin{equation}
    \mathrm{BIC}(M)
    =
    -2\ell_M(\hat\theta_M)
    +
    q_M\log(n_{\mathrm{train}}),
    \label{eq:bic-section5}
\end{equation}
where \(\ell_M(\hat\theta_M)\) is the maximized log-likelihood for candidate model \(M\), \(q_M\) is the number of estimated parameters, and \(n_{\mathrm{train}}\) is the number of training anchors. The Bayesian information criterion is used as a likelihood-based model selection rule with an explicit penalty for model dimension \citep{schwarz1978bic}. The participant random intercept is retained throughout the selection procedure.

For held-out test participants, prediction uses the marginal fixed-effect component,
\begin{equation}
    \hat y_{ij}^{\mathrm{stat}}
    =
    \hat\beta_0
    +
    x_{ij}^{\top}\hat\beta.
    \label{eq:lmm-fixed-prediction-section5}
\end{equation}
The participant-specific random intercept is not used for test prediction because the held-out participants are not observed during training. Equivalently, the prediction uses \(E(b_{0i})=0\) for new participants. This rule keeps the evaluation aligned with the participant-level split and avoids using subject-specific outcome information from the test set.

\paragraph{GRU-D.}
The second comparator is GRU-D, a recurrent model designed for multivariate time series with missing values and irregular observation times \citep{che2018grud}. It is included because the present data contain both missingness and variable-specific elapsed time since last observation.

Let \(x_t\in\mathbb{R}^{D}\) denote the observed feature vector at time \(t\), \(m_t\in\{0,1\}^{D}\) the observation mask, and \(\delta_t\in\mathbb{R}_{+}^{D}\) the elapsed time since each variable was last observed. GRU-D learns variable-specific input decay and hidden-state decay terms,
\begin{align}
    \gamma_t^x
    &=
    \exp\!\left\{
    -\max\left(0, W_\gamma^x\delta_t+b_\gamma^x\right)
    \right\}, \label{eq:grud-input-decay-section5}\\
    \gamma_t^h
    &=
    \exp\!\left\{
    -\max\left(0, W_\gamma^h\delta_t+b_\gamma^h\right)
    \right\}. \label{eq:grud-hidden-decay-section5}
\end{align}
The input decay is used to form a completed input vector,
\begin{equation}
    \tilde x_t
    =
    m_t\odot x_t
    +
    (1-m_t)\odot
    \left[
        \gamma_t^x\odot x_t^{\mathrm{last}}
        +
        (1-\gamma_t^x)\odot \bar x
    \right],
    \label{eq:grud-imputed-input-section5}
\end{equation}
where \(x_t^{\mathrm{last}}\) is the most recent observed value for each variable, \(\bar x\) is the empirical mean vector computed from the training data, and \(\odot\) denotes elementwise multiplication. The hidden state is decayed as
\begin{equation}
    \tilde h_{t-1}
    =
    \gamma_t^h\odot h_{t-1}.
    \label{eq:grud-hidden-decay-applied-section5}
\end{equation}
The recurrent update then follows a gated recurrent unit form using the completed input, the decayed hidden state, and the observation mask:
\begin{align}
    r_t
    &=
    \sigma(W_r\tilde x_t+U_r\tilde h_{t-1}+V_r m_t+b_r),\\
    q_t
    &=
    \sigma(W_q\tilde x_t+U_q\tilde h_{t-1}+V_q m_t+b_q),\\
    \tilde h_t^{\,*}
    &=
    \tanh(W_h\tilde x_t+U_h(r_t\odot \tilde h_{t-1})+V_hm_t+b_h),\\
    h_t
    &=
    (1-q_t)\odot \tilde h_{t-1}
    +
    q_t\odot \tilde h_t^{\,*}.
    \label{eq:grud-update-section5}
\end{align}
The final hidden state is passed to a regression head to predict 24-month CDR-SB change. GRU-D is a strong non-transformer comparator because its architecture directly encodes two features of the study data: stale measurements and informative missingness.

\paragraph{STraTS.}
The third comparator is STraTS, a transformer-based model for irregularly sampled multivariate clinical time series \citep{tipirneni2021strats}. STraTS is included because it uses observation-level triplets rather than forcing all variables onto a dense visit-by-variable grid.

For a history represented by observed triplets
\[
    H_i=\{(\tau_{ij},k_{ij},v_{ij})\}_{j=1}^{m_i},
\]
STraTS maps each triplet to an embedding
\begin{equation}
    e_{ij}
    =
    e_{\tau}(\tau_{ij})
    +
    e_{k}(k_{ij})
    +
    e_{v}(v_{ij}),
    \label{eq:strats-embedding-section5}
\end{equation}
where \(e_{\tau}\), \(e_k\), and \(e_v\) encode observation time, variable identity, and observed value. The embedded sequence is then processed by self-attention. In a standard attention head, the pre-softmax score is
\begin{equation}
    s_{iab}
    =
    \frac{
    \langle W_Qe_{ia},W_Ke_{ib}\rangle
    }{\sqrt{d_h}},
    \label{eq:strats-attention-score-section5}
\end{equation}
and the attention weight is
\begin{equation}
    \alpha_{iab}
    =
    \frac{\exp(s_{iab})}
    {\sum_{c=1}^{m_i}\exp(s_{iac})}.
    \label{eq:strats-attention-weight-section5}
\end{equation}
The resulting token representations are pooled and passed to a regression head. This comparator is important because it isolates the value of the proposed residualization and gap-aware attention beyond the use of a transformer-family architecture itself. The proposed model shares the triplet-style longitudinal representation with STraTS, but adds the mixed-effects reference term and a learned time-gap penalty inside the attention score.

\paragraph{Proposed residual gap-aware transformer.}
The proposed model is the structured statistical-neural model defined in Section~\ref{sec:proposed-method}. It first fits the mixed-effects reference model on the training partition, then learns a transformer-based residual from the pre-anchor longitudinal clinical and biomarker history. In the experiment, this model is evaluated against the three comparators above under the same participant-level splits and the same test metrics.

\subsection{Participant-level splitting and repeated-seed design}

All experiments use participant-level splitting. This choice is essential because a single participant can contribute more than one eligible MCI anchor. A visit-level split would allow earlier and later anchors from the same participant to appear in different data partitions, which would make the test set partially identifiable from the training set. The split is therefore performed at the participant level before any anchor-level model fitting or validation step is carried out.

For each random seed, 80\% of participants are assigned to the training and validation pool, and the remaining 20\% are held out as the test set. Validation participants are selected only from the training pool. After the participant split is fixed, all eligible anchors from a participant stay in the same partition. Thus, every training, validation, and test set contains complete participant-level anchor histories rather than isolated visits. The repeated-seed analysis uses five participant-level random seeds: 42, 43, 44, 45, and 46.

This design gives two advantages. First, it protects the test evaluation from within-subject leakage. Second, it reduces the dependence of the final results on a single random split. The reported performance summaries are therefore based on repeated participant-level train--validation--test partitions rather than on one arbitrary partition of the cohort.

\subsection{Hyperparameters and implementation}

Implementation details are summarized in Table~\ref{tab:implementation-settings}. The linear mixed-effects model is used as the longitudinal statistical reference. GRU-D and STraTS are used as neural comparators for irregular clinical time series, and the proposed model is evaluated using the architecture defined in Section~\ref{sec:proposed-method}. All model fitting, feature selection, and early stopping decisions are carried out within the training and validation partitions only.

\begin{table}[H]
    \centering
    \caption{Implementation settings for the compared model families.}
    \label{tab:implementation-settings}
    \small
    \setlength{\tabcolsep}{6pt}
    \renewcommand{\arraystretch}{1.25}
    \begin{tabular}{p{0.22\linewidth}p{0.26\linewidth}p{0.42\linewidth}}
        \toprule
        Model family & Role in the comparison & Main implementation settings \\
        \midrule
        Linear mixed-effects model &
        Longitudinal statistical reference &
        Participant-level random intercept; fixed-effect terms selected by Bayesian information criterion within the training partition; fixed-effect prediction used for held-out test participants. \\
        \addlinespace[0.25em]
        GRU-D &
        Missingness-aware recurrent comparator &
        Hidden size 48; dropout 0.2; batch size 64; learning rate \(10^{-3}\); early stopping based on validation performance. \\
        \addlinespace[0.25em]
        STraTS &
        Observation-level transformer comparator &
        Hidden dimension 48; two transformer layers; four attention heads; dropout 0.2; attention dropout 0.1; batch size 64; learning rate \(3\times 10^{-4}\); early stopping based on validation performance. \\
        \addlinespace[0.25em]
        Proposed residual gap-aware transformer &
        Structured statistical-neural model &
        Architecture defined in Section~\ref{sec:proposed-method}; hidden dimension 64; two transformer layers; four attention heads; dropout 0.1; attention dropout 0.1; batch size 64; initial gap-penalty scale 0.1 per year; early stopping based on validation performance. \\
        \bottomrule
    \end{tabular}
\end{table}

The settings are kept moderate across the neural models. The purpose of the experiment is to compare model families under the same longitudinal progression task, rather than to perform an exhaustive architecture search. The proposed model is therefore evaluated with a small transformer encoder and compared against recurrent and transformer baselines trained under the same participant-level split logic.

\subsection{Evaluation metrics}

Evaluation is performed on held-out test anchors from held-out participants. Let \(\mathcal{I}_{\mathrm{test}}\) denote the test anchor set and let
\[
    N_{\mathrm{test}}=|\mathcal{I}_{\mathrm{test}}|.
\]
For each test anchor \((i,j)\), let \(y_{ij}\) be the observed 24-month CDR-SB change and \(\hat y_{ij}\) be the model prediction. The prediction error is
\[
    e_{ij}=y_{ij}-\hat y_{ij}.
\]
We report mean squared error,
\begin{equation}
    \mathrm{MSE}
    =
    \frac{1}{N_{\mathrm{test}}}
    \sum_{(i,j)\in\mathcal{I}_{\mathrm{test}}}
    e_{ij}^{2},
    \label{eq:mse-section5}
\end{equation}
mean absolute error,
\begin{equation}
    \mathrm{MAE}
    =
    \frac{1}{N_{\mathrm{test}}}
    \sum_{(i,j)\in\mathcal{I}_{\mathrm{test}}}
    |e_{ij}|,
    \label{eq:mae-section5}
\end{equation}
root mean squared error,
\begin{equation}
    \mathrm{RMSE}
    =
    \sqrt{\mathrm{MSE}},
    \label{eq:rmse-section5}
\end{equation}
and Pearson prediction--observation correlation,
\begin{equation}
    \mathrm{Corr}
    =
    \frac{
    \sum_{(i,j)\in\mathcal{I}_{\mathrm{test}}}
    (y_{ij}-\bar y_{\mathrm{test}})
    (\hat y_{ij}-\bar{\hat y}_{\mathrm{test}})
    }
    {
    \sqrt{
    \sum_{(i,j)\in\mathcal{I}_{\mathrm{test}}}
    (y_{ij}-\bar y_{\mathrm{test}})^2
    }
    \sqrt{
    \sum_{(i,j)\in\mathcal{I}_{\mathrm{test}}}
    (\hat y_{ij}-\bar{\hat y}_{\mathrm{test}})^2
    }
    },
    \label{eq:corr-section5}
\end{equation}
where
\[
    \bar y_{\mathrm{test}}
    =
    \frac{1}{N_{\mathrm{test}}}
    \sum_{(i,j)\in\mathcal{I}_{\mathrm{test}}}
    y_{ij},
    \qquad
    \bar{\hat y}_{\mathrm{test}}
    =
    \frac{1}{N_{\mathrm{test}}}
    \sum_{(i,j)\in\mathcal{I}_{\mathrm{test}}}
    \hat y_{ij}.
\]
The error metrics measure numerical accuracy of predicted CDR-SB change, while the correlation measures how well the model preserves the ordering of test anchors by progression magnitude.

For each metric, repeated-seed results are summarized by the mean and an approximate 95\% split-level interval across the five participant-level random seeds. If \(M_s\) denotes the metric value for seed \(s\in\{1,\dots,S\}\), with \(S=5\), then
\[
    \bar M
    =
    \frac{1}{S}\sum_{s=1}^{S}M_s,
    \qquad
    \mathrm{SE}(\bar M)
    =
    \frac{1}{\sqrt{S}}
    \left\{
    \frac{1}{S-1}
    \sum_{s=1}^{S}(M_s-\bar M)^2
    \right\}^{1/2}.
\]
The reported interval is
\begin{equation}
    \bar M
    \pm
    t_{S-1,0.975}\,\mathrm{SE}(\bar M),
    \label{eq:seed-ci-section5}
\end{equation}
where \(t_{S-1,0.975}\) is the 97.5th percentile of the \(t\) distribution with \(S-1\) degrees of freedom. These intervals are used as descriptive summaries of split-level stability, rather than as formal sampling-based confidence intervals.

\section{Results}

\subsection{Distribution of the 24-month CDR-SB-change target}

\begin{figure}[t]
    \centering
    \includegraphics[width=0.88\linewidth]{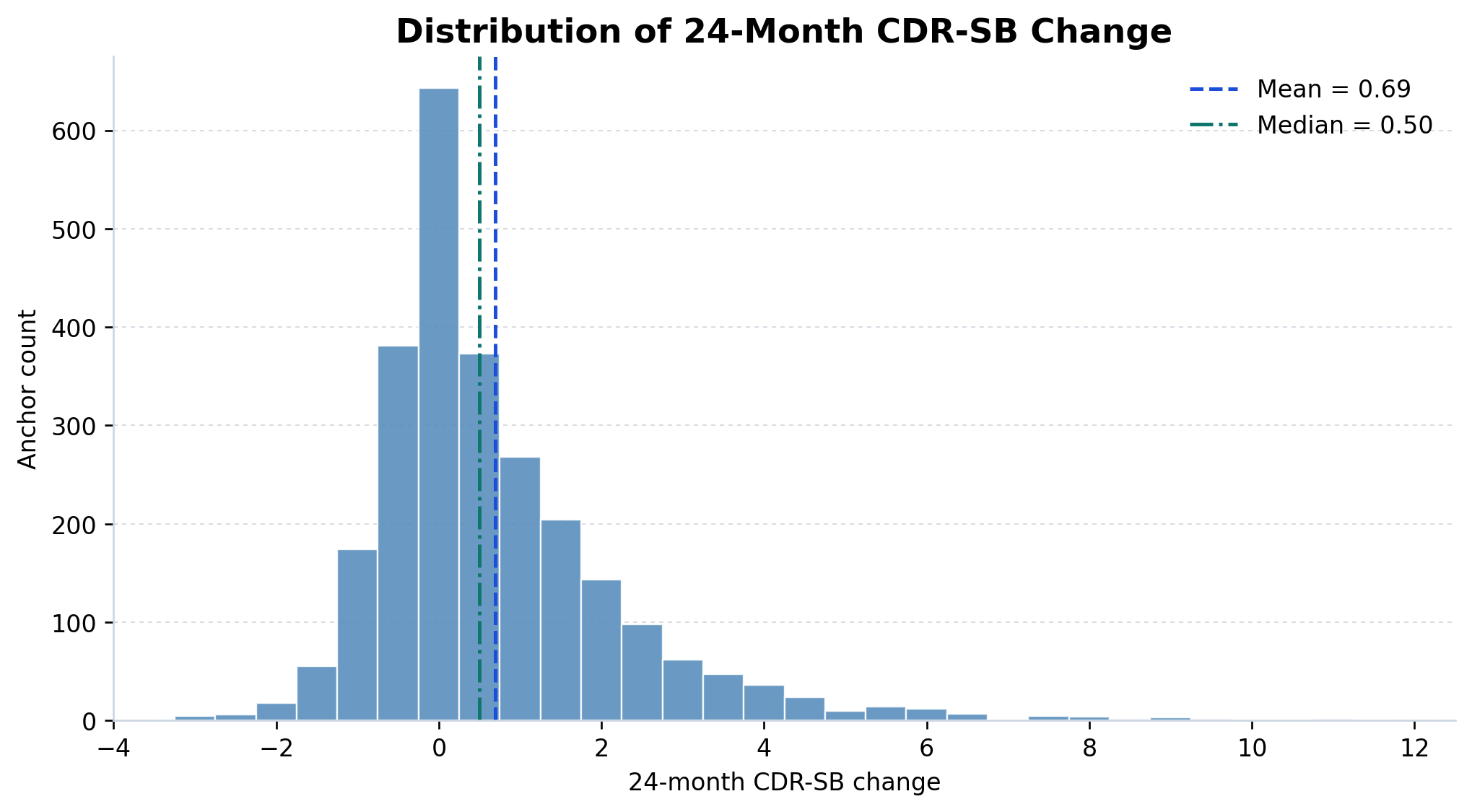}
    \caption{Distribution of 24-month CDR-SB change in the primary analysis cohort. The distribution is centered near mild worsening and has a right tail, indicating that the analysis contains both common small changes and less frequent larger worsening events.}
    \label{fig:delta-distribution}
\end{figure}

We first examine the empirical distribution of the response variable. Figure~\ref{fig:delta-distribution} shows the distribution of 24-month CDR-SB change across the labeled anchors used in the primary analysis. The outcome is centered near mild worsening, with mean 0.69 and median 0.50, and the distribution has a clear right tail. This shape is clinically meaningful. Many MCI anchors show small changes over the follow-up interval, while a smaller group shows larger worsening. A useful model therefore has to predict both common mild changes and less frequent larger increases in CDR-SB.

The response distribution also helps explain why the task is more demanding than predicting raw future severity. A model that mainly carries forward current disease burden can perform well on raw future CDR-SB, but 24-month CDR-SB change asks the model to predict subsequent worsening after the anchor. The distribution in Figure~\ref{fig:delta-distribution} therefore motivates the use of both error metrics and correlation: error metrics evaluate numerical closeness to observed change, while correlation evaluates whether the model preserves relative ordering across smaller and larger worsening patterns.

\subsection{Primary performance comparison}

Table~\ref{tab:continuous-results} reports repeated-seed test performance across the four compared model families under participant-level splitting. The proposed residual gap-aware transformer achieves the best mean performance across all four criteria: mean squared error, mean absolute error, root mean squared error, and prediction--observation correlation. The mixed-effects baseline provides the repeated-measures statistical reference, GRU-D provides a strong missingness-aware recurrent comparator, and STraTS provides the observation-level transformer comparator. Within this comparison set, the proposed method gives the strongest mean performance profile.

\begin{table}[H]
    \centering
    \caption{Repeated-seed test performance on 24-month CDR-SB change. Values are mean $\pm$ approximate 95\% split-level interval over participant-level seeds 42, 43, 44, 45, and 46. Lower is better for error metrics; higher is better for correlation.}
    \label{tab:continuous-results}
    \small
    \begin{tabular}{lcccc}
        \toprule
        Method & MSE & MAE & RMSE & Corr \\
        \midrule
        Mixed-effects baseline & 2.227 $\pm$ 0.311 & 1.095 $\pm$ 0.020 & 1.488 $\pm$ 0.104 & 0.382 $\pm$ 0.020 \\
        GRU-D & 2.030 $\pm$ 0.172 & 0.985 $\pm$ 0.013 & 1.424 $\pm$ 0.060 & 0.455 $\pm$ 0.035 \\
        STraTS & 2.100 $\pm$ 0.226 & 1.007 $\pm$ 0.012 & 1.447 $\pm$ 0.078 & 0.413 $\pm$ 0.044 \\
        Proposed residual gap-aware transformer & \textbf{1.936 $\pm$ 0.193} & \textbf{0.970 $\pm$ 0.015} & \textbf{1.389 $\pm$ 0.069} & \textbf{0.483 $\pm$ 0.050} \\
        \bottomrule
    \end{tabular}
\end{table}

\begin{figure}[t]
    \centering
    \includegraphics[width=0.95\linewidth]{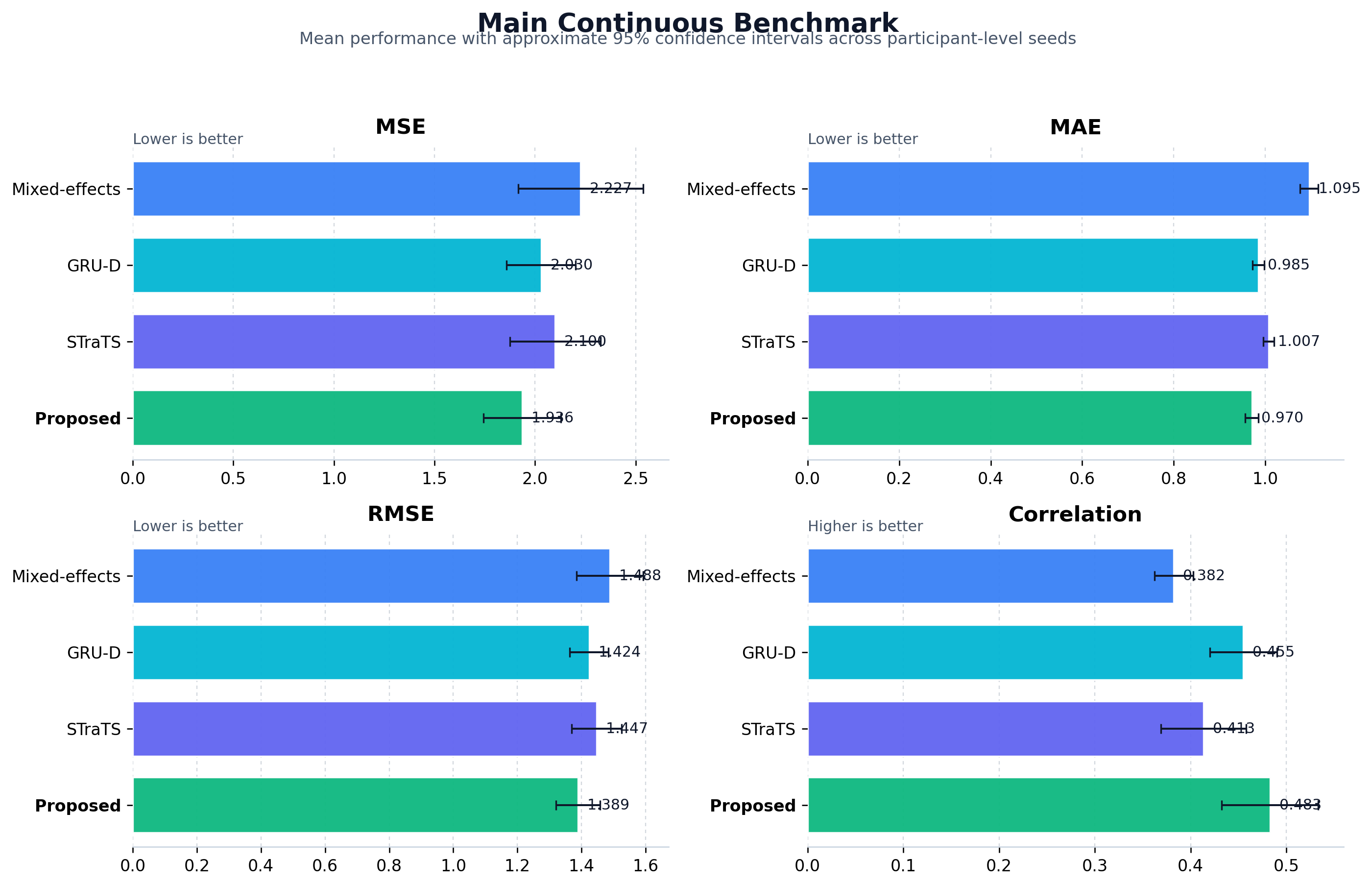}
    \caption{Main performance comparison for the 24-month CDR-SB-change task. Bars show repeated-seed mean performance, with approximate 95\% split-level intervals across participant-level seeds. The proposed residual gap-aware transformer gives the best mean result across all four metrics.}
    \label{fig:main-metric-overview}
\end{figure}

Figure~\ref{fig:main-metric-overview} gives the same comparison visually. The proposed method has the lowest mean error on MSE, MAE, and RMSE, and it has the highest mean correlation. This visual summary is useful because it shows that the proposed method improves both numerical accuracy and agreement with the observed ordering of progression magnitude.

Several aspects of the comparison are important. The mixed-effects baseline is a genuine longitudinal statistical comparator because it accounts for repeated anchors through participant random intercepts during training. GRU-D is also a strong baseline because it directly models missingness and elapsed time in irregular clinical sequences. STraTS is the closest transformer-family comparator because it uses observation-level triplets. The proposed model improves over all three, which supports the value of combining the mixed-effects reference, residual learning, and gap-aware attention in a single model.

\subsection{Magnitude of improvement}

Table~\ref{tab:relative-gains} summarizes the relative improvement of the proposed residual gap-aware transformer over each competing model family. Relative to the mixed-effects baseline, the proposed method reduces mean squared error by 13.1\%, mean absolute error by 11.4\%, and root mean squared error by 6.7\%, while increasing prediction--observation correlation by 26.4\%. These are the largest relative gains because the mixed-effects model is a strong but lower-flexibility statistical reference.

Relative to GRU-D, the proposed method still improves every metric, although the margins are smaller. This is important because GRU-D is specifically designed for irregular time series with missing values. Relative to STraTS, the proposed method again improves every metric, which shows that the advantage comes from more than using a transformer architecture. The improvement over STraTS is especially relevant because both STraTS and the proposed model use observation-level triplet representations of irregular longitudinal histories.

\begin{table}[t]
    \centering
    \caption{Relative improvement of the proposed residual gap-aware transformer over competing model families, computed from the repeated-seed means in Table~\ref{tab:continuous-results}. Negative values indicate lower error; positive values indicate higher correlation.}
    \label{tab:relative-gains}
    \small
    \begin{tabular}{lcccc}
        \toprule
        Comparison target & $\Delta$MSE & $\Delta$MAE & $\Delta$RMSE & $\Delta$Corr \\
        \midrule
        Versus mixed-effects baseline & $-0.291$ ($-13.1\%$) & $-0.125$ ($-11.4\%$) & $-0.099$ ($-6.7\%$) & $+0.101$ ($+26.4\%$) \\
        Versus GRU-D & $-0.094$ ($-4.6\%$) & $-0.015$ ($-1.5\%$) & $-0.035$ ($-2.5\%$) & $+0.028$ ($+6.2\%$) \\
        Versus STraTS & $-0.164$ ($-7.8\%$) & $-0.037$ ($-3.7\%$) & $-0.058$ ($-4.0\%$) & $+0.070$ ($+17.0\%$) \\
        \bottomrule
    \end{tabular}
\end{table}

These results give a direct interpretation of the contribution. The proposed method improves substantially over the mixed-effects statistical reference, which indicates that the pre-anchor longitudinal history contains residual signal beyond the fixed-effect statistical prediction. It also improves over GRU-D, a missingness-aware recurrent model, and over STraTS, an observation-level transformer model. The pattern therefore supports the combined structure of the proposed method: statistical anchoring first, residual sequence learning second, and time-gap-aware attention inside the residual learner.

\subsection{Repeated-seed stability}

\begin{figure}[t]
    \centering
    \includegraphics[width=0.95\linewidth]{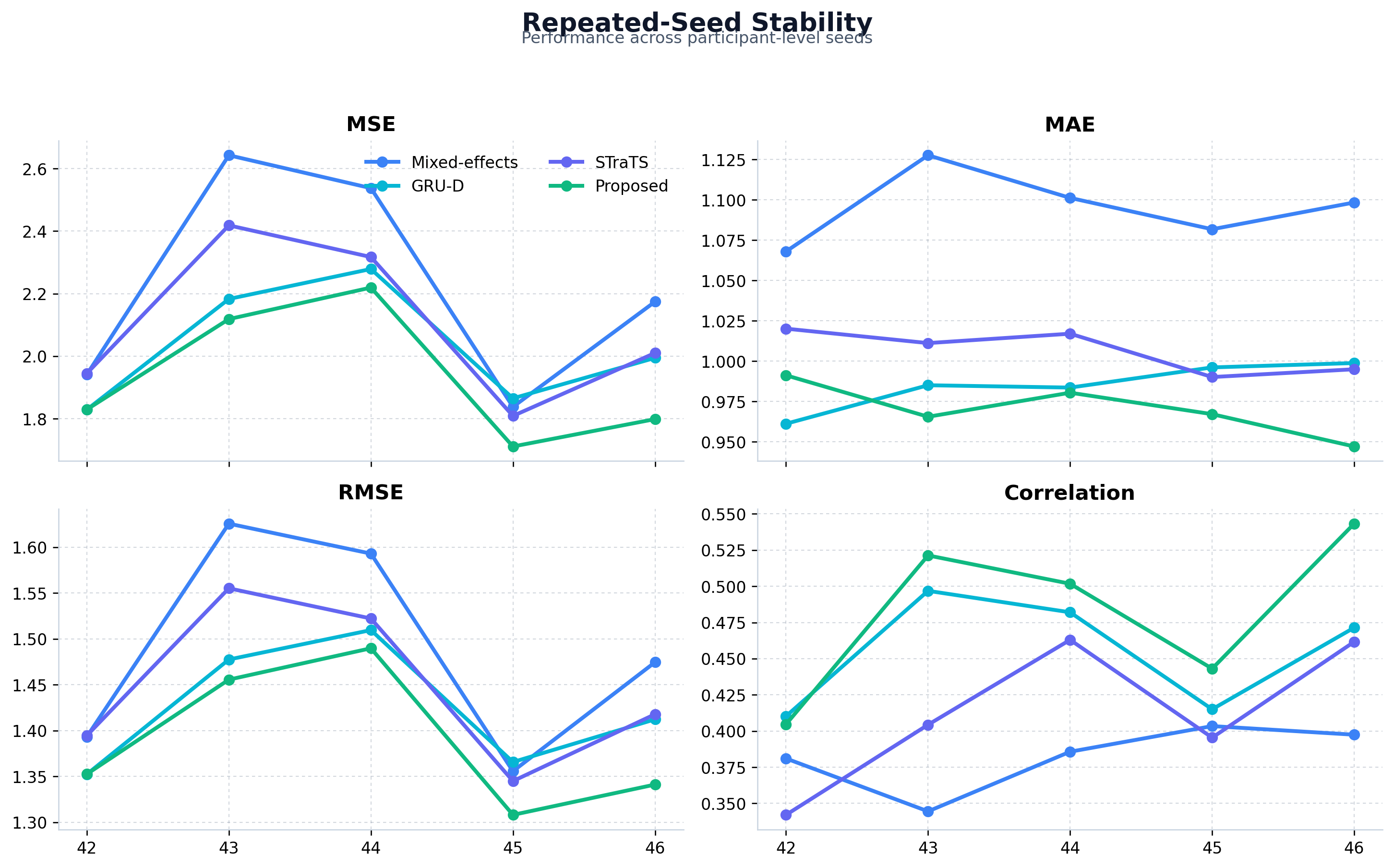}
    \caption{Repeated-seed stability across participant-level splits. Each panel shows test performance across seeds 42, 43, 44, 45, and 46. The proposed residual gap-aware transformer remains consistently competitive and has the best mean performance across the reported metrics.}
    \label{fig:seed-stability}
\end{figure}

The repeated-seed analysis examines whether the performance advantage is stable across participant-level splits. Figure~\ref{fig:seed-stability} displays the metric values for seeds 42, 43, 44, 45, and 46. The proposed method is consistently competitive across all seeds and is usually the best or among the best methods for each metric. Its strongest pattern appears in correlation, where it is highest across most seeds and highest on average.

The stability plot also shows that the analysis itself varies across participant splits. For example, the absolute error levels change across seeds, especially for MSE and RMSE. This variation is expected in a participant-level split with repeated anchors and heterogeneous progression patterns. The key point is that the proposed method maintains a strong position despite this split-to-split variability. This supports the conclusion that the improvement in Table~\ref{tab:continuous-results} is not driven by a single favorable partition.

\subsection{Main empirical message: statistical anchoring and gap-aware residual learning}

The main empirical message is that the proposed model gains performance by combining two complementary sources of structure. The mixed-effects reference provides a statistical anchor for repeated-anchor longitudinal data. It uses anchor-level covariates and participant random intercepts during training to represent the repeated-measures structure of the cohort. The transformer residual then focuses on the part of 24-month CDR-SB change that remains after this statistical reference has been fitted. This division of labor gives the neural component a clearer role than direct end-to-end prediction from the full history.

The improvement over the mixed-effects baseline shows that the longitudinal clinical and biomarker history contains predictive information beyond the fixed-effect statistical prediction. This is the first important finding. The mixed-effects model already accounts for repeated anchors and uses the same anchor-level clinical and biomarker summaries. The residual gain therefore suggests that the pre-anchor trajectory itself, including timing and irregular observation patterns, adds useful information for medium-horizon progression prediction.

The improvement over GRU-D gives a second message. GRU-D is a strong comparator for irregular clinical time series because it explicitly models missingness and elapsed time. The proposed method achieves a better repeated-seed mean profile while using a different strategy: it represents the history as observation-level triplets and lets attention weights depend directly on temporal distance. This indicates that gap-aware attention is a useful way to use irregular timing in the present CDR-SB-change task.

The improvement over STraTS gives a third message. STraTS is the closest transformer-family comparator because it also uses observation-level triplets. The proposed method adds two elements on top of this representation: residualization against a mixed-effects reference and a learned nonnegative time-gap penalty in attention. The gain over STraTS therefore supports the proposed architecture as more than a standard transformer applied to the same inputs.

Taken together, the results support the central claim of the paper: medium-horizon Alzheimer's disease progression prediction benefits from aligning the statistical reference, neural residual learner, and time representation with the longitudinal data structure. The 24-month CDR-SB-change response focuses the task on worsening after the MCI anchor. Participant-level splitting protects the evaluation from within-subject leakage. The mixed-effects component supplies the longitudinal statistical reference. The gap-aware transformer then learns residual trajectory information from irregular clinical and biomarker histories. This combination gives the proposed model its empirical advantage.

\section{Discussion}

This study shows that medium-horizon Alzheimer's disease progression can be modeled more effectively when the prediction target, covariate history, statistical reference, and neural architecture are matched to the same longitudinal data structure. The main empirical result is that the proposed residual gap-aware transformer achieves the best repeated-seed mean performance across MSE, MAE, RMSE, and prediction--observation correlation for 24-month CDR-SB change. This result is informative because the comparison includes a mixed-effects repeated-measures baseline, a missingness-aware recurrent neural model, and an observation-level transformer model for irregular clinical time series. The improvement therefore reflects more than generic model flexibility.

The first implication concerns the endpoint. The 24-month CDR-SB-change response places the prediction target on worsening after the anchor visit. This differs from predicting raw future CDR-SB, which remains strongly tied to current CDR-SB in the present cohort. The change-score formulation makes the task more directly about medium-horizon progression magnitude. Because CDR-SB is an ordered composite clinical scale, the response should be interpreted as a quantitative change score rather than as a strictly continuous biological measurement. This interpretation supports regression modeling while remaining faithful to the clinical nature of the outcome.

The second implication concerns longitudinal clinical and biomarker history construction. The cohort profile shows that same-visit structural magnetic resonance imaging and cerebrospinal fluid measurements are limited at the anchor visit, while any-prior availability is broader. The study therefore uses historically observed biomarker information through prior summaries and recency variables. This design better reflects the way longitudinal Alzheimer's disease data are collected. It also avoids restricting the analysis to a narrow subset with strict same-visit biomarker coverage.

The third implication concerns the role of the mixed-effects reference model. The proposed model is built as a residual learner relative to a longitudinal statistical baseline. This gives the neural component a clear role: it models trajectory information beyond the anchor-level fixed-effect structure captured by the mixed-effects model. The empirical gain over the mixed-effects baseline suggests that such residual sequence information is useful. The gain over STraTS further suggests that the advantage comes from the combination of residualization and time-gap-aware attention rather than from using a transformer alone.

The comparison with GRU-D is also informative. GRU-D is designed for irregular time series with missing values, and it remains a strong comparator in this study. The proposed model improves over GRU-D on the repeated-seed mean of all reported metrics, but the margin is smaller than the gain over the mixed-effects baseline. This pattern is scientifically useful because it shows that missingness-aware recurrent modeling is already well matched to this data structure, while the proposed model adds value by combining time-aware representation learning with an explicit statistical reference.

The results therefore support a specific methodological conclusion. For irregular longitudinal clinical and biomarker histories in Alzheimer's disease, a useful modeling strategy is to align four elements: a clinically interpretable medium-horizon response, participant-level evaluation, a longitudinal statistical comparator, and a neural architecture that directly represents irregular timing. Under this alignment, the proposed residual gap-aware transformer gives the strongest repeated-seed mean performance in the present ADNI-based analysis.

\section{Limitations}

Several limitations should guide the interpretation of the results. First, the analysis is based on ADNI. ADNI is deeply characterized and carefully curated, but it is still a research cohort. Its participants, measurement schedule, and data quality differ from many routine-care settings. The present results therefore establish performance within an ADNI-derived progression analysis. Broader clinical transportability requires temporal validation, external validation, or internal--external validation across independent cohorts \citep{collins2015tripod,steyerberg2016validation}.

Second, the primary response is based on CDR-SB change. CDR-SB is clinically meaningful and widely used, but it remains an ordered composite score with bounded values and scale-specific measurement properties. Treating 24-month CDR-SB change as a quantitative regression response is reasonable for the present prediction task, yet future work should also examine robustness to alternative outcome definitions, including raw future CDR-SB, clinically meaningful worsening thresholds, diagnosis conversion, and longer-horizon change.

Third, the biomarker inputs are table-derived summaries rather than raw imaging or raw molecular data. This choice keeps the paper focused on longitudinal prognosis from harmonized clinical and biomarker histories, but it leaves open the question of whether raw magnetic resonance imaging, positron emission tomography, or learned image representations could add signal beyond the regional summaries used here. This is especially relevant because modern Alzheimer's disease research increasingly frames disease status through biological markers of amyloid, tau, and neurodegeneration \citep{jack2018niaaa}.

Fourth, the biomarker-history construction uses most recent prior biomarker summaries and modality-recency variables. This is a practical and interpretable way to retain historical information under irregular follow-up, but it remains a summary of the full biomarker trajectory. More flexible approaches could model biomarker evolution directly, represent uncertainty in stale measurements, or combine longitudinal biomarker trajectories with clinical endpoints in a joint modeling framework.

Fifth, the repeated-seed analysis uses five participant-level random seeds. This improves over a single split and shows that the proposed model remains strong across multiple partitions, but the intervals across seeds should be interpreted as split-level stability summaries rather than full sampling uncertainty. Larger repeated-split studies, bootstrap evaluation, or external validation would give a stronger assessment of stability.

Finally, the theoretical results in this paper are structural properties of the proposed empirical objective and attention mechanism. They clarify the residual-risk equivalence, the recovery of the statistical reference as a feasible model, temporal attenuation from the nonnegative gap penalty, and Lipschitz stability under boundedness conditions. These results help explain the architecture, but they do not provide a full finite-sample generalization theorem for the learned deep model.

\section{Conclusion and Future Work}

This paper develops a medium-horizon Alzheimer's disease progression analysis centered on 24-month CDR-SB change from longitudinal clinical and biomarker ADNI histories. The proposed residual gap-aware transformer combines a mixed-effects statistical reference, observation-level longitudinal tokenization, and a learned nonnegative time-gap penalty inside attention. Under participant-level repeated-seed evaluation, the proposed model achieves the strongest mean performance across all reported metrics and improves over a mixed-effects baseline, GRU-D, and STraTS.

The main conclusion is that the value of the proposed method comes from the alignment between the clinical prediction problem and the model structure. The response is a quantitative change score over a clinically interpretable two-year horizon. The inputs preserve irregular longitudinal histories rather than forcing strict same-visit biomarker coverage. The baseline comparison includes a repeated-measures statistical reference. The proposed model then learns residual trajectory signal with attention weights that account for temporal distance. This combination gives a clear contribution to longitudinal Alzheimer's disease progression modeling.

Future work should first test external validity. The next stage should evaluate the model across additional Alzheimer's disease cohorts, temporal ADNI splits, and more routine-care datasets. This direction is especially important for statistical-neural longitudinal modeling, because our recent Parkinson's disease study on longitudinal voice biomarkers found that neural flexibility in small clinical cohorts requires careful validation against interpretable statistical references \citep{tong2026parkinson}. More broadly, clinical prediction models require validation beyond the development sample before their performance can be interpreted as transportable \citep{collins2015tripod,steyerberg2016validation}.

A second direction is calibration and clinical utility. The present paper reports error metrics and prediction--observation correlation, which are appropriate for a regression analysis of CDR-SB change. Future work should also assess calibration of predicted progression magnitude, calibration within risk or progression strata, and clinical decision value. Calibration is a central requirement for trustworthy prediction models \citep{vancalster2019calibration}, and decision-curve analysis can help assess whether predicted progression adds value for clinically relevant decisions \citep{vickers2006decision}.

A third direction is modality expansion. The current analysis uses structured magnetic resonance imaging summaries and cerebrospinal fluid biomarkers. Future models could incorporate raw imaging, positron emission tomography, learned image embeddings, or richer biomarker trajectories. This would connect the statistical-neural progression model more closely to biological staging frameworks for Alzheimer's disease \citep{jack2018niaaa}. Such extensions should be evaluated carefully because richer modalities can increase predictive signal while also increasing missingness, computational burden, and risk of overfitting.

A fourth direction is endpoint sensitivity. The 24-month CDR-SB-change endpoint is a useful primary response for the present paper, but the broader scientific question includes multiple horizons and multiple forms of progression. Future studies should compare 12-month, 24-month, and 36-month change; raw future CDR-SB; diagnosis conversion; and time-to-progression outcomes. These analyses would clarify whether the proposed residual gap-aware structure is specifically strongest for quantitative change or whether it also improves other clinically relevant endpoints.

Finally, future work should improve interpretability of the learned residual component. The mixed-effects reference already provides an interpretable statistical component. The next step is to characterize which time points, variables, and data sources contribute most to the residual transformer prediction. Attention patterns, source-specific ablations, and counterfactual removal of historical biomarkers could help explain whether the model gains mainly from cognitive trajectories, biomarker recency, structural neurodegeneration summaries, cerebrospinal fluid history, or their interactions over time.

Overall, the present study supports a focused conclusion: structured statistical-neural modeling can improve 24-month CDR-SB-change prediction from longitudinal clinical and biomarker ADNI histories when the outcome, covariate history, baseline model, and evaluation protocol are defined consistently. Future work should test the scope, clinical reliability, and biological interpretation of that finding.

\newpage
\appendix

\section{Proofs of Structural Properties}
\label{app:proofs}

This appendix gives the proofs of Propositions~\ref{prop:residual-equivalence}--\ref{prop:stability}. The results are deterministic properties of the empirical objective and the proposed model map under the structural assumptions stated in Section~\ref{sec:proposed-method}.

\subsection{Proof of Proposition~\ref{prop:residual-equivalence}}

For fixed statistical reference predictions, define
\[
u_i=y_i-g_i^{\mathrm{stat}}.
\]
For every \(\theta\in\Theta\) and every training anchor \(i\),
\[
y_i-g_i^{\mathrm{stat}}-r_\theta(H_i)
=
u_i-r_\theta(H_i).
\]
Squaring both sides gives
\[
\left(y_i-g_i^{\mathrm{stat}}-r_\theta(H_i)\right)^2
=
\left(u_i-r_\theta(H_i)\right)^2.
\]
Averaging over \(i=1,\dots,n\) yields
\[
\frac{1}{n}
\sum_{i=1}^n
\left(y_i-g_i^{\mathrm{stat}}-r_\theta(H_i)\right)^2
=
\frac{1}{n}
\sum_{i=1}^n
\left(u_i-r_\theta(H_i)\right)^2.
\]
Thus the original empirical objective and the residual empirical objective are identical for every \(\theta\). Therefore the two optimization problems have the same objective values, the same infimum, and the same set of minimizers. This proves Proposition~\ref{prop:residual-equivalence}. \hfill \(\square\)

\subsection{Proof of Proposition~\ref{prop:no-worse}}

By Assumption~\ref{assump:zero}, there exists \(\theta_0\in\Theta\) such that
\[
r_{\theta_0}(H)=0
\]
for every admissible history \(H\). In particular,
\[
r_{\theta_0}(H_i)=0
\]
for every training history \(H_i\). Evaluating the empirical loss at this feasible parameter value gives
\[
\mathcal{L}_n(\theta_0)
=
\frac{1}{n}
\sum_{i=1}^n
\left(y_i-g_i^{\mathrm{stat}}-r_{\theta_0}(H_i)\right)^2.
\]
Using \(r_{\theta_0}(H_i)=0\),
\[
\mathcal{L}_n(\theta_0)
=
\frac{1}{n}
\sum_{i=1}^n
\left(y_i-g_i^{\mathrm{stat}}\right)^2.
\]
Since the infimum over \(\Theta\) is no larger than the value of the objective at any feasible parameter point,
\[
\inf_{\theta\in\Theta}\mathcal{L}_n(\theta)
\le
\mathcal{L}_n(\theta_0)
=
\frac{1}{n}
\sum_{i=1}^n
\left(y_i-g_i^{\mathrm{stat}}\right)^2.
\]
This proves Proposition~\ref{prop:no-worse}. \hfill \(\square\)

\subsection{Proof of Proposition~\ref{prop:gap-score}}

Fix a layer \(\ell\), a head \(h\), a query token \(a\), and a key token \(b\). Holding the content term fixed, write
\[
s_{ab}^{(\ell,h)}(d)
=
c_{ab}^{(\ell,h)}-\lambda_{\ell,h}d,
\qquad
d\ge 0.
\]
By Assumption~\ref{assump:gap},
\[
\lambda_{\ell,h}\ge 0.
\]
For any \(d_2>d_1\ge 0\),
\[
s_{ab}^{(\ell,h)}(d_2)-s_{ab}^{(\ell,h)}(d_1)
=
-\lambda_{\ell,h}(d_2-d_1).
\]
Because \(d_2-d_1>0\) and \(\lambda_{\ell,h}\ge 0\),
\[
s_{ab}^{(\ell,h)}(d_2)-s_{ab}^{(\ell,h)}(d_1)\le 0.
\]
Hence \(s_{ab}^{(\ell,h)}(d)\) is nonincreasing in \(d\). If \(\lambda_{\ell,h}>0\), then
\[
-\lambda_{\ell,h}(d_2-d_1)<0
\]
for every \(d_2>d_1\), so the score is strictly decreasing. This proves Proposition~\ref{prop:gap-score}. \hfill \(\square\)

\subsection{Proof of Proposition~\ref{prop:gap-softmax}}

Fix a query token, a head, and all key scores except the score of key token \(b\). Let
\[
d=|\tau_a-\tau_b|
\]
and write the score of token \(b\) as
\[
s_b(d)=c_b-\lambda d,
\]
where \(c_b\) is fixed and \(\lambda=\lambda_{\ell,h}>0\). Let
\[
A=\sum_{c\neq b}\exp(s_c),
\]
where each \(s_c\) for \(c\neq b\) is fixed. Define
\[
z(d)=\exp(c_b-\lambda d).
\]
The softmax weight of token \(b\) is then
\[
\alpha_b(d)
=
\frac{z(d)}{A+z(d)}.
\]
Since
\[
z'(d)
=
-\lambda \exp(c_b-\lambda d)
=
-\lambda z(d),
\]
differentiating \(\alpha_b(d)\) gives
\[
\alpha_b'(d)
=
\frac{z'(d)(A+z(d))-z(d)z'(d)}{(A+z(d))^2}.
\]
The terms involving \(z(d)z'(d)\) cancel, so
\[
\alpha_b'(d)
=
\frac{A z'(d)}{(A+z(d))^2}.
\]
Substituting \(z'(d)=-\lambda z(d)\) gives
\[
\alpha_b'(d)
=
-\lambda
\frac{A z(d)}{(A+z(d))^2}.
\]
Because \(\lambda>0\), \(A\ge 0\), and \(z(d)>0\),
\[
\alpha_b'(d)\le 0.
\]
Thus the attention weight is nonincreasing in \(d\). If token \(b\) competes with at least one other key token, then \(A>0\), and therefore
\[
\alpha_b'(d)<0.
\]
The decrease is then strict. This proves Proposition~\ref{prop:gap-softmax}. \hfill \(\square\)

\subsection{Proof of Proposition~\ref{prop:stability}}

Let \(H,H'\in\mathcal{H}\) be two admissible histories. By Assumption~\ref{assump:lipschitz}, each history has at most \(m_{\max}<\infty\) observed tokens, and observed times and values lie in bounded sets after preprocessing. Variable identities are equipped with the discrete metric. Therefore each admissible history can be represented, after padding or another fixed-length embedding convention, in a bounded finite-dimensional domain.

By Assumption~\ref{assump:lipschitz}, the tokenization map is Lipschitz on \(\mathcal{H}\). Hence there exists a finite constant \(L_0\) such that
\[
\|Z^{(0)}(H)-Z^{(0)}(H')\|
\le
L_0 d_{\mathcal{H}}(H,H').
\]

For a fixed layer \(\ell\) and head \(h\), the query, key, and value maps are linear:
\[
Q=ZW_Q^{(\ell,h)},\qquad
K=ZW_K^{(\ell,h)},\qquad
V=ZW_V^{(\ell,h)}.
\]
The corresponding weight matrices have finite operator norms by Assumption~\ref{assump:lipschitz}; hence these maps are Lipschitz on the admissible domain. The content score map
\[
(Q,K)\mapsto QK^\top/\sqrt{d_h}
\]
is bilinear and is Lipschitz on bounded domains. The temporal penalty
\[
(\tau_a,\tau_b)\mapsto \lambda_{\ell,h}|\tau_a-\tau_b|
\]
is Lipschitz in the time coordinates for finite \(\lambda_{\ell,h}\). Therefore the complete attention score map is Lipschitz on the admissible domain.

The row-wise softmax map is Lipschitz on any fixed finite-dimensional bounded score set. For a score vector \(s\in\mathbb{R}^m\), its Jacobian has entries
\[
\frac{\partial \alpha_j}{\partial s_k}
=
\alpha_j(\mathbf{1}\{j=k\}-\alpha_k).
\]
Since \(0\le \alpha_j\le 1\), the Jacobian entries are bounded. Because \(m\le m_{\max}\), the corresponding operator norm is uniformly bounded over admissible histories. Thus the softmax map is Lipschitz on the score sets generated by the model.

The attention head output is
\[
O^{(\ell,h)}=\alpha^{(\ell,h)}V^{(\ell,h)}.
\]
On the bounded admissible domain, this output is a product of bounded Lipschitz maps and is therefore Lipschitz. Concatenation over finitely many heads and multiplication by \(W_O^{(\ell)}\) preserve Lipschitz continuity. Residual addition and the feed-forward update are Lipschitz by Assumption~\ref{assump:lipschitz}. Hence each transformer layer defines a Lipschitz map
\[
Z^{(\ell-1)}\mapsto Z^{(\ell)}.
\]
Because the encoder has finitely many layers, the full encoder map
\[
Z^{(0)}\mapsto Z^{(L)}
\]
is Lipschitz with some finite constant \(L_{\mathrm{enc}}\).

The pooling score
\[
z\mapsto w_p^\top\tanh(W_pz+b_p)
\]
is a composition of Lipschitz maps and is therefore Lipschitz. The pooling softmax is Lipschitz by the bounded-Jacobian argument above, again using the fact that the number of tokens is bounded by \(m_{\max}\). The weighted sum
\[
h=\sum_j \pi_j z_j
\]
is Lipschitz on the bounded admissible domain because both the weights \(\pi_j\) and the encoded tokens \(z_j\) are bounded Lipschitz functions of the input history. Finally, the residual head
\[
h\mapsto w_r^\top h+b_r
\]
is linear and hence Lipschitz.

Combining the Lipschitz constants from tokenization, the encoder, pooling, and residual head, there exists a finite constant
\[
L_\theta
=
L_{\mathrm{head}}L_{\mathrm{pool}}L_{\mathrm{enc}}L_0
\]
such that
\[
|r_\theta(H)-r_\theta(H')|
\le
L_\theta d_{\mathcal{H}}(H,H').
\]
This proves Proposition~\ref{prop:stability}. \hfill \(\square\)

\end{document}